\documentclass{article} %
\usepackage{iclr2026_conference,times}

\usepackage{amsmath,amsfonts,bm}

\def\eqref#1{equation~\ref{#1}}

\def\1{\bm{1}}

\DeclareMathAlphabet{\mathsfit}{\encodingdefault}{\sfdefault}{m}{sl}
\SetMathAlphabet{\mathsfit}{bold}{\encodingdefault}{\sfdefault}{bx}{n}

\usepackage{hyperref}
\usepackage{url}

\title{Surprisingly Simple and Effective Multi-Domain Graph Foundation Model through Graph-to-Table Alignment}

\author{Chunyu Hu$^{1}$, Tianyin Liao$^{1}$, Ge Lan$^{1}$, 
        Xingxuan Zhang$^{3}$, Jianxin Li$^{2}$, Peng Cui$^{3}$, Ziwei Zhang$^{2}$\thanks{Corresponding author.}\\
        $^{1}$Nankai University, 
        $^{2}$Beihang University, 
        $^{3}$Tsinghua University\\
\texttt{huchunyu@mail.nankai.edu.cn, 1120230329@mail.nankai.edu.cn} \\
\texttt{lange@nankai.edu.cn, xingxuanzhang@hotmail.com, lijx@buaa.edu.cn} \\
\texttt{cuip@tsinghua.edu.cn, zwzhang@buaa.edu.cn} 
}

\usepackage{caption}
\usepackage{subcaption}
\usepackage{booktabs}
\usepackage{amsmath}
\usepackage{amssymb}
\usepackage{mathtools}
\usepackage{amsthm}
\newcommand{\name}{GTAlign}
\usepackage{bm}
\usepackage{multicol}
\usepackage{multirow}
\usepackage{enumitem}
\newcommand{\thirdstage}{In-Context Graph Adaptation}
\usepackage{subcaption}
\usepackage[table]{xcolor}

\iclrfinalcopy %
\begin{document}

\maketitle

\begin{abstract}
Graph Foundation Models (GFMs) have emerged as a promising paradigm for learning transferable representations across diverse graph domains. Recent advancements in GFMs have been largely dominated by two paradigms: Graph Neural Network and Large Language Model (LLM) based methods. However, these methods often face a fundamental dilemma between training with limited data and a heavy reliance on textual attributes. Tabular foundation models (TFMs) offer a potential alternative, as node features and representations can be naturally organized in a tabular form. However, how to enable TFMs to effectively capture structural information of graphs remains largely unexplored. The key challenge is to learn a graph-to-table alignment mechanism that enables graph structural understanding for TFMs. To address this, we propose \textbf{\name}, a surprisingly simple yet effective \underline{G}raph-to-\underline{T}able \underline{Align}ment framework for text-free Graph Foundation Model. Specifically, we first pretrain a graph encoder that maps diverse graphs into a unified latent space to capture domain-agnostic graph representations. To further bridge the gap between graph topology and the tabular representation space, we propose community-guided continual pre-training, where pseudo-labels derived from graph community are used to construct few-shot prediction episodes. Lastly, we adapt the graph encoder for an unseen target domain and perform in-context inference. Extensive experiments on five benchmark datasets demonstrate that \name~significantly outperforms state-of-the-art baselines on both node and graph classification, offering a simple, effective, and text-free GFM model. Code will be released upon acceptance.

\end{abstract}

\section{Introduction}
 
Graph Neural Networks (GNNs) have been the cornerstone of graph representation learning~\cite{DBLP:conf/iclr/KipfW17, velivckovic2018graph, xu2018how}, boasting a wide range of applications. However, traditional GNNs are typically task-specific and domain-dependent, requiring training from scratch on specific domains and failing to generalize across unseen domains~\cite{hustrategies}. 

Inspired by the success of Large Language Models (LLMs) as foundation models in natural language processing~\cite{zhou2025comprehensive}, recent research attention has been shifting towards Graph Foundation Models (GFMs)~\cite{mao2024position, zi2024prog, yu2025samgpt}, which aim to learn general-purpose graph representations through pre-training on multi-domain graph datasets, which can be adapted for diverse downstream tasks~\cite{wang2025graph}.
Existing GFMs can be categorized into three types based on their backbone architectures~\cite{liu2025graph}: GNN-based~\cite{liu2023graphprompt}, LLM-based~\cite{wang2024instructgraph}, and GNN+LLM-based models~\cite{liu2024can}. %
However, despite the initial success, these architectures still face notable limitations. Particularly, GNN-based methods are often constrained by the scale and diversity of available graph pre-training corpora, which limits their ability to generalize across different domains and limits their performance in downstream tasks~\cite{liu2025graph}. Meanwhile, LLM-based and GNN+LLM-based models often rely heavily on textual attributes, which restricts the model’s usability and effectiveness~\cite{eremeevturning}.

These limitations motivate us to explore whether Tabular Foundation Models (TFMs) can provide a potential paradigm for text-free GFMs. TFMs, such as TabPFNv2~\cite{hollmann2025accurate} and LimiX~\cite{zhang2025limix}, have demonstrated a remarkable ability to generalize across unseen distributions for tabular prediction tasks~\cite{hollmanntabpfn, hollmann2025accurate, zhang2025limix, grinsztajn2025tabpfn}. These models are built upon the Prior-data Fitted Network (PFN) framework and pretrained on large-scale and diverse synthetic data, enabling powerful in-context learning capabilities. However, adapting TFMs for graph data processing is non-trivial. Although node features can be naturally organized in a tabular form, enabling TFMs to effectively understand graph structures remains largely unexplored, particularly for the multi-domain settings. The difficulty lies in the fact that graph structural information is not merely encoded in individual node attributes, but also expressed through structured of different scales such as neighborhoods and communities, which differ greatly in different domains. In contrast, TFMs are solely pretrained to model row-column relationships in tabular data, and thus lack the inherent ability to understand different domain graph structures. Therefore, how to learn a graph-to-table alignment mechanism that enables TFMs to effectively exploit graph structural information remains a key challenge.

To address this challenge, we propose \textbf{\name}, a \underline{S}imple yet \underline{E}ffective graph-to-\underline{T}able alignment framework for text-free \underline{G}raph \underline{F}oundation \underline{M}odel. Specifically, we first pre-train a graph encoder on multi-domain graph datasets to map graph data into a unified latent space, producing domain-agnostic and topology-aware graph representations. Individual nodes or graphs are encoded into unified graph representations, which are treated as the tabular tokens and fed as inputs for the TFM. Second, to bridge the gap between graph topology and the tabular representation space of the TFM, we introduce a community-guided continual pre-training. We derive pseudo-labels from graph community structures and use them to construct few-shot prediction episodes. By jointly training the graph encoder and the TFM on these episodes, the TFM learns how to exploit topology-aware graph representations for conditional prediction. Lastly, for an unseen target domain, we adapt the graph encoder using a few labeled examples and perform in-context inference with a frozen TFM. Extensive experiments on five benchmark datasets show that \name~achieves strong few-shot performance on both node and graph classification.

Our main contributions are summarized as follows:
\begin{itemize}[leftmargin=0.5cm]
    \item We explore text-free GFM from a graph-to-table alignment perspective by representing graph-structured data as tabular instances and leveraging pretrained TFMs for prediction. To the best of our knowledge, this is the first work that adapts TFMs for multi-domain GFMs.
    
    \item We propose \name, a simple and effective graph-to-table alignment framework. \name~first pre-trains a universal graph encoder on multi-domain graph data, then introduces a community-guided continual pre-training to align representations, and finally adapts the encoder with in-context inference.
    
    \item We conduct extensive experiments on five benchmark datasets for both node and graph classification. The results demonstrate that \name~achieves extremely strong few-shot performance compared with state-of-the-art baselines.
\end{itemize}

\begin{figure*}[ht]
  \begin{center}
    \centerline{\includegraphics[width=0.98\textwidth]{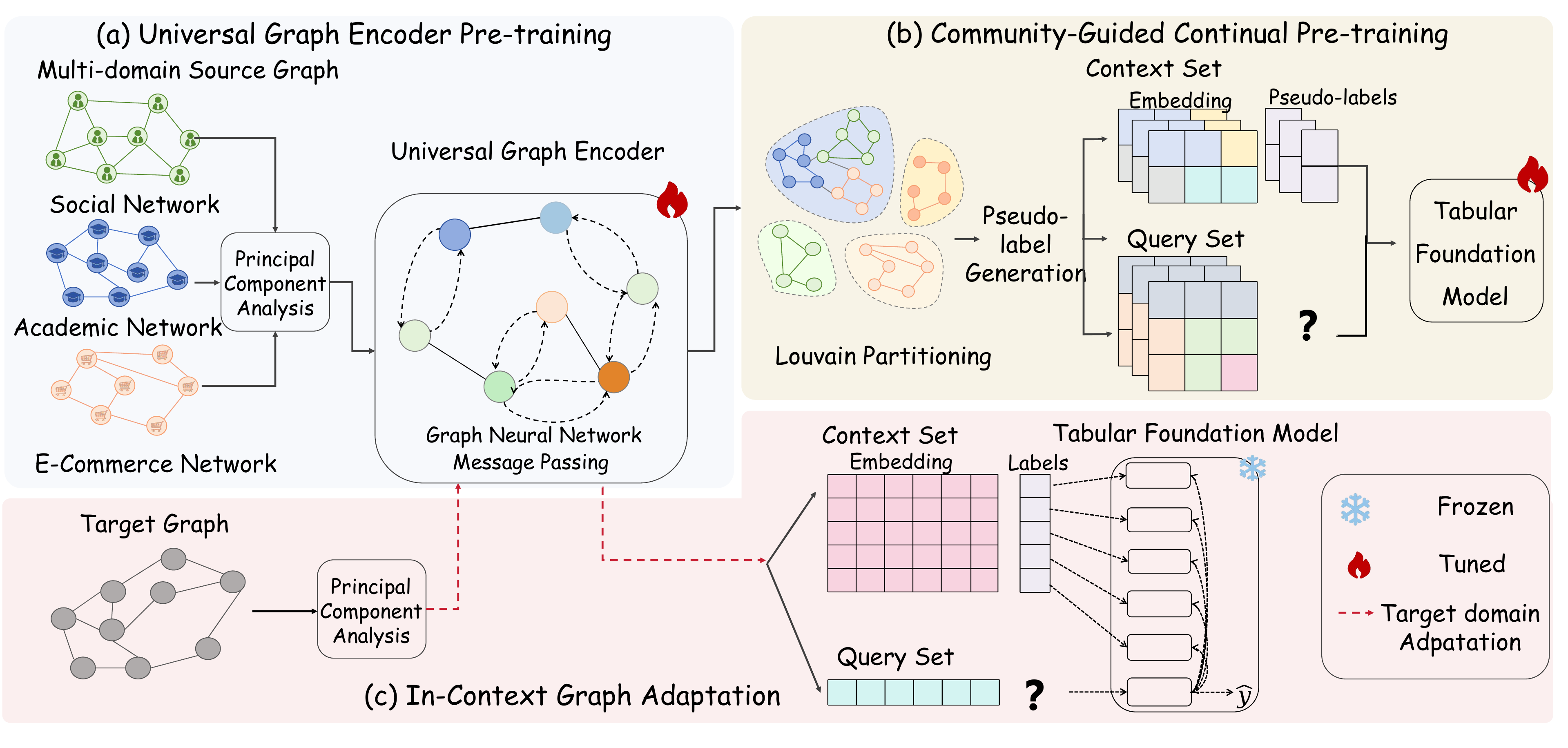}}
    \caption{An overview of the \name~framework. It consists of three stages: (a) Universal Graph Encoder Pre-training: a graph encoder is pre-trained via contrastive learning on multi-domain graphs. (b) Community-Guided Continual Pre-training, where Louvain pseudo-labels construct few-shot in-context episodes to align graph representations with the TFM. (c) \thirdstage: on an unseen target domain, the graph encoder is adapted with few labeled target samples and the TFM performs in-context learning to make downstream predictions.}
    \label{fig:framework}
  \end{center}
  \vspace{-0.3cm}
\end{figure*}

\section{Method}
In this section, we introduce \name. The overall framework is illustrated in Fig~\ref{fig:framework}. First, we introduce the notations. Then, we elaborate the three stages of our method: (1) Universal Graph Encoder Pre-training; (2) Community-Guided Continual Pre-training; (3) \thirdstage.

\subsection{Notations}
\paragraph{Graph.} We define a graph as $ G = (\mathcal{V}, \mathcal{E}, \mathbf{X})$, where $\mathcal{V}$
represents the set of nodes, and $\mathcal{E}$ denotes the set of edges. $\mathbf{X} \in \mathbb{R}^{|\mathcal{V}| \times d}$ is the node feature matrix and $d$ is the dimensionality of the feature space. Let $A\in\{0,1\}^{N\times N}$ denote the adjacency matrix of $G$, where $A_{ij}=1$ if $(v_i,v_j)\in E$, and $A_{ij}=0$ otherwise. A collection of graphs is denoted as $\mathcal{G} = \{G_1, G_2, \dots, G_m\}$.

\paragraph{Cross-domain Graph Foundation Model.} A Graph Foundation Model is pretrained on a diverse set of source domains $\mathcal{D}_S = \{\mathcal{G}_{S_{1}}, \mathcal{G}_{S_{2}}, \dots\}$. The objective is to learn generalizable structural and statistical representations that can be adapted to a target domain $\mathcal{D}_T$. We specifically consider the cross-domain setting, where $\mathcal{D}_T \cap \mathcal{D}_S = \emptyset$, meaning the target domain remains unseen during the pre-training phase.

\paragraph{Tabular Foundation Model.}
A tabular dataset is represented as $\mathcal{T} = (\mathbf{D}, \mathbf{y})$, where $\mathbf{D} \in \mathbb{R}^{m \times k}$ contains $m$ rows and $k$ columns. The label vector $\mathbf{y}$ provides the supervision for a downstream task. A TFM is a pre-trained model $f_\Theta$ that learns a universal inference mechanism. Leveraging the framework of PFNs enables the model to perform in-context learning, where it produces the posterior predictive distribution for a query instance $\mathbf{X}_{q}$ given a context set of labeled examples.
\begin{equation}
    p(\mathbf{y}_q|\mathbf{X}_q,\mathbf{X}_{\text{ctx}},\mathbf{y}_{\text{ctx}})\approx f_\Theta(\mathbf{X}_q,\mathbf{X}_{\text{ctx}},\mathbf{y}_{\text{ctx}}),
\end{equation}
where $( \mathbf{X}_{\text{ctx}}, \mathbf{y}_{\text{ctx}})$ are provided as in-context samples. 

In this work, we adapt TFMs for GFMs by treating each node or graph as a table row and encoding them into representations $\mathbf{H}$. Consequently, the TFM inference for a target query set $\mathbf{H}_q$, given a context set $\mathcal{S}_{\text{ctx}} = \{(\mathbf{H}_{\text{ctx}}, \mathbf{y}_{\text{ctx}})\}$, is formulated as:
\begin{equation}
    \hat{\mathbf{y}_{q}} = f_\Theta(\mathbf{H}_q, \mathbf{H}_{\text{ctx}}, \mathbf{y}_{\text{ctx}}),
\label{eq: tfm_gfm}
\end{equation}
where $\mathbf{H}_{\text{ctx}}$ and $\mathbf{y}_{\text{ctx}}$ are the concatenated embeddings and labels of the context samples, respectively. Next, we introduce how to obtain the representation $\mathbf{H}$ while incorporating structural information through a three-stage model.
 
\subsection{Universal Graph Encoder Pre-training}
The primary objective of the first stage is to train a universal graph encoder $E_\phi$ capable of extracting general structural representations across diverse source domains $\mathcal{D}_S$. A fundamental challenge in multi-domain pre-training is the discrepancy in node feature dimensions. Specifically, let $\mathbf{X}_i$ and $\mathbf{X}_j$ represent the raw node feature matrices from two distinct source domains $\mathcal{G}_{S_i}$ and $\mathcal{G}_{S_j}$ respectively. Their feature dimensionalities do not necessarily match, i.e., $d_i \neq d_j$. To address this, we follow previous feature alignment methods~\cite{yu2025samgpt, zhao2024all} and utilize Principal Component Analysis (PCA) to map diverse raw feature matrix into a unified dimensionality $\tilde{d}$:
\begin{equation}
    \tilde{\mathbf{X}} = \text{PCA}(\mathbf{X}) \in \mathbb{R}^{|\mathcal{V}| \times \tilde{d}},
\end{equation}
where $\tilde{\mathbf{X}}$ serves as the standardized input for the encoder across all datasets. We employ graph contrastive learning for learning unsupervised representations of graph data~\cite{you2020graph}. For a given graph $G$, we generate two correlated augmented views $\hat{G}^{(1)}$ and $\hat{G}^{(2)}$ by applying an edge perturbation strategy. The graph encoder $E_\phi$ processes these views to produce node-level embeddings, which are then aggregated by a readout function $\mathcal{R}$ into graph-level representations:
\begin{equation}\label{eq:readout}
    \mathbf{h}^{(1)} = \mathcal{R}(E_\phi(\tilde{\mathbf{X}}, \mathbf{A}^{(1)}_{\text{aug}})),
    \mathbf{h}^{(2)} = \mathcal{R}(E_\phi(\tilde{\mathbf{X}}, \mathbf{A}^{(2)}_{\text{aug}})).
\end{equation}

To optimize the universal graph encoder, we employ a contrastive loss $\mathcal{L}_{\text{pre}}$ that maximizes the cosine similarity between positive pairs while minimizing it for negative samples. Formally, $\mathcal{L}_{\text{pre}}$ is defined over a set of graphs $\mathcal{G}$ as:
\begin{equation}
    \mathcal{L}_{\text{pre}} =- \sum_{G \in \mathcal{G}} \log \frac{
        \exp\bigl(\text{sim}(\mathbf{h}^{(1)}, \mathbf{h}^{(2)}) / \tau
    )}{
        \sum_{G' \in \mathcal{G}} \exp\bigl(\text{sim}(\mathbf{h}^{(1)}, \mathbf{h}'^{(2)}) / \tau
    )},
\end{equation}
where $\text{sim}(\cdot, \cdot)$ denotes the cosine similarity metric and $\tau$ is a temperature scaling factor. For a representation $\mathbf{h}_i^{(1)}$, the corresponding positive sample is $\mathbf{h}_i^{(2)}$, which is derived from the same graph instance $G_i$ but through a different stochastic augmentation view. The negative samples $\mathbf{h'}_j^{(2)} (j \neq i)$ are the representations obtained from other distinct graphs. 

\subsection{Community-Guided Continual Pre-training}
TFMs are pre-trained on massive synthetic tabular datasets. %
However, graph-structured data is characterized by non-Euclidean topologies, whose statistical properties differ fundamentally from the generic tabular spaces familiar to TFMs. Therefore, how to align the tabular feature space and graph topology space remains a key bottleneck for building tabular-based GFMs.

To bridge this gap, we introduce community-guided continual pre-training, which aligns topology-aware graph representations with the tabular representation space. Specifically, we construct pseudo-labels for nodes by partitioning each source graph into communities, and use these pseudo-labels to continually train the universal graph encoder $E_\phi$ and the TFM $f_\Theta$ jointly as follows.

\paragraph{Community-based Pseudo-Label Generation.}
For each source graph $G \in \mathcal{D}_S$, let $\mathbf{A}$ be its adjacency matrix. We apply Louvain community detection~\cite{blondel2008fast} to obtain pseudo-labels for nodes. Louvain partitions the graph by maximizing the resolution-aware modularity:
\begin{equation}
    Q_\rho = \frac{1}{2M}\sum_{i,j}
    \left(A_{ij}-\rho\frac{k_i k_j}{2M}\right)
    \mathbb{I}[c_i=c_j],
\end{equation}
where $A_{ij}$ denotes the adjacency between nodes $v_i$ and $v_j$, $k_i$ is the degree of node $v_i$, $M=\frac{1}{2}\sum_i k_i$ is the number of edges, $c_i$ is the community assignment of $v_i$, $\mathbb{I}[c_i=c_j]$ is an indicator function that equals 1 if nodes $v_i$ and $v_j$ are assigned to the same community, and 0 otherwise, and $\rho$ is the resolution parameter controlling the granularity of the partition. At each continual pre-training step, we sample the resolution parameter $\rho \sim \mathrm{Uniform}(\rho_{\min}, \rho_{\max})$. A larger $\rho$ encourages finer partitions with more communities, while a smaller $\rho$ leads to coarser communities. This multi-resolution design exposes the model to structural patterns at different granularities. The detected communities provide a topology-derived supervision signal. We assign each node its community index as a pseudo-label
$    \tilde{y}_i = c_i, v_i \in \mathcal{V} $ 
and denote the pseudo-label vector over all nodes by $\tilde{\mathbf{y}}$.

\paragraph{Few-Shot Continual Pre-training.}

Given the community-based pseudo-labels, we construct few-shot in-context learning episodes to continually align the graph encoder with the TFM. Specifically, we utilize the graph encoder $E_\phi$ to extract node representations:
\begin{equation}
    \mathbf{H} = E_\phi(\tilde{\mathbf{X}}, \mathbf{A}),
    \label{eq:node_representations}
\end{equation}
where $\mathbf{h}_i = \mathbf{H}_{i,:}$ denotes the representation of node $v_i$. Given the pseudo-labels $\tilde{\mathbf{y}}$, we construct a few-shot in-context learning episode. For each pseudo-class, we sample $m$ nodes as context samples, forming the context node set $\mathcal{V}_{\mathrm{ctx}}$:
\begin{equation}
    \mathcal{S}_{\mathrm{ctx}} =
    \{(\mathbf{h}_i,\tilde{y}_i) \mid v_i \in \mathcal{V}_{\mathrm{ctx}}\}.
\end{equation}

A randomly sampled subset of the remaining nodes are used as query samples:
\begin{equation}
    \mathcal{S}_{q} =
    \{(\mathbf{h}_j,\tilde{y}_j) \mid v_j \in \mathcal{V}_{q}\}.
\end{equation}

Let $\mathbf{H}_{\mathrm{ctx}}$ and $\tilde{\mathbf{y}}_{\mathrm{ctx}}$ denote the context embeddings and their pseudo-labels, and let $\mathbf{H}_{q}$ and $\tilde{\mathbf{y}}_{q}$ denote the query embeddings and their pseudo-labels. The TFM predicts the pseudo-labels of query nodes conditioned on the context set:
\begin{equation}
    \hat{\mathbf{y}}_{q}
    =
    f_\Theta(
    \mathbf{H}_{q},
    \mathbf{H}_{\mathrm{ctx}},
    \tilde{\mathbf{y}}_{\mathrm{ctx}}
    ).
\end{equation}

To align the GNN with the TFM, the joint system is optimized by minimizing the cross-entropy loss over the query set predictions. The alignment loss $\mathcal{L}_{align}$ is formulated as:
\begin{equation}
    \mathcal{L}_{\mathrm{align}}
    =
    -
    \sum \nolimits_{v_j \in \mathcal{V}_{q}}
    \log
    P
    \left(
    \tilde{y}_j
    \mid
    \mathbf{H}_{q},
    \mathbf{H}_{\mathrm{ctx}},
    \tilde{\mathbf{y}}_{\mathrm{ctx}}
    \right),
\end{equation}
where $P(\cdot)$ denotes the posterior predictive distribution produced by the TFM, and $\tilde{y}_j$ is the pseudo-label of query node $v_j$. By optimizing this objective, the graph encoder learns to produce topology-aware representations that are compatible with the tabular representation space of the TFM, while the TFM learns how to exploit such representations for conditional prediction. %
For graph-level tasks, graph representations are obtained by applying the same encoder followed by the readout function defined in Eq.~\ref{eq:readout}. Therefore, both node- and graph-level representations can be processed by the same aligned TFM for downstream few-shot prediction.

\subsection{\thirdstage}

After universal graph encoder pre-training and community-guided continual pre-training, we adapt the graph encoder to an unseen target domain $\mathcal{D}_T$, while keeping the TFM frozen. This stage aims to make the graph representations more task-relevant for the target domain without updating the TFM.

We use $\mathbf{h}_i$ to denote the instance representation fed into the TFM. For node classification on a target graph, the graph encoder produces node representations as Eq.~\ref{eq:node_representations}, the representation of node $v_i$ is given by $\mathbf{h}_i = \mathbf{H}_{i,:}$. For graph classification, given a target graph, the encoder first produces node representations, and the graph-level representation is obtained by the readout function $\mathbf{h}_i = R(\mathbf{H}_i)$.

Consistent with existing methods~\cite{yuan2025much, yu2025samgpt}, we first fine-tune the graph encoder $E_\phi$ on a few labeled samples $\mathcal{S}_{\mathrm{ctx}}=\{(\mathbf{h}_i,y_i)\}_{i=1}^{k}$ from the target domain to better capture target-specific structural patterns. Specifically, we adopt a prototypical metric learning objective. For the context set, we compute the prototype vector for each class $c$ as the mean of the embeddings of its observed samples:
\begin{equation}
\mathbf{c}_c = \frac{1}{|\mathcal{S}_{\mathrm{ctx}}^c|} \sum_{(\mathbf{h}_i,y_i)\in \mathcal{S}_{\mathrm{ctx}}^c} \mathbf{h}_i,
\end{equation}
where $\mathcal{S}_{\mathrm{ctx}}^c$ denotes the subset of context samples belonging to class $c$. The probability that a labeled sample belongs to class $c$ is computed by a softmax over cosine similarities:
\begin{equation}
p(y_i=c\mid \mathbf{h}_i)=
\frac{\exp(\mathrm{sim}(\mathbf{h}_i,\mathbf{c}_c))}
{\sum_{c'}\exp(\mathrm{sim}(\mathbf{h}_i,\mathbf{c}_{c'}))}.
\end{equation}
The fine-tuning objective is the cross-entropy loss over the labeled context samples:
\begin{equation}
\mathcal{L}_{\mathrm{ft}} =
-\sum \nolimits_{(\mathbf{h}_i,y_i)\in \mathcal{S}_{\mathrm{ctx}}}
\log p(y_i\mid \mathbf{h}_i).
\end{equation}

After fine-tuning, we recompute the context and query representations from the adapted graph encoder. For a query graph $G_q$, we feed its embedding $\mathbf{h}_q$ together with the context set embeddings $\mathbf{H}_{\mathrm{ctx}} = [\mathbf{h}_1 \dots \mathbf{h}_k]^\top$ and corresponding labels $\mathbf{y}_{\mathrm{ctx}}$ into the TFM $f_\Theta$ to produce the final prediction through in-context learning as
\begin{equation}
    \hat{y}_q = f_\Theta(\mathbf{h}_q, \mathbf{H}_{\mathrm{ctx}}, \mathbf{y}_{\mathrm{ctx}}).
\end{equation}

\section{Experiments}

\begin{table*}[!t]
\caption{The results (Accuracy ± standard deviation for one hundred runs, $\%$) of five-shot node classification. The best results are shown in \textbf{bold} and the runner-ups are \underline{underlined}.}
\label{tab:five_shot_node_cls}
\vspace{-0.3cm}
\begin{center}
\resizebox{\textwidth}{!}{
\begin{tabular}{lccccc}
\toprule
Model / Target  & Cora & CiteSeer & PubMed & Photo & Computers \\
\midrule
\rowcolor{gray!15}
\multicolumn{6}{c}{\textit{End-to-end GNN}}\\
 GCN~\cite{DBLP:conf/iclr/KipfW17} & 46.53 ± 5.36 & 46.59 ± 6.31 & 51.07 ± 5.89 & 56.17 ± 5.75 & 51.82 ± 6.93 \\
 GAT~\cite{velivckovic2018graph} & 46.47 ± 7.10 & 46.09 ± 5.91 & 50.47 ± 7.26 & 53.11 ± 7.80 & 49.78 ± 5.68 \\
\midrule
\rowcolor{gray!15}
\multicolumn{6}{c}{\textit{Graph Self-Supervised Pre-training}}\\
GCC~\cite{qiu2020gcc} & 47.05 ± 7.88 & 45.12 ± 6.78 & 52.45 ± 7.67 & 59.76 ± 7.15 & 51.96 ± 5.31 \\
DGI~\cite{velivckovicdeep} & 47.77 ± 6.85 & 47.22 ± 5.66 & 52.43 ± 5.67 & 62.60 ± 6.41 & 54.66 ± 7.03 \\
GraphCL~\cite{you2020graph} & 49.61 ± 6.67 & 48.76 ± 6.72 & 52.81 ± 5.27 & 59.55 ± 7.50 & 57.88 ± 7.36 \\
DSSL~\cite{xiao2022decoupled} & 46.28 ± 7.61 & 48.44 ± 7.26 & 52.42 ± 7.57 & 62.35 ± 6.94 & 54.15 ± 7.58 \\
GraphACL~\cite{xiao2023simple} & 51.38 ± 7.06 & 49.71 ± 7.12 & 54.48 ± 6.01 & 65.87 ± 6.54 & 59.72 ± 5.36 \\
\midrule
\rowcolor{gray!15}
\multicolumn{6}{c}{\textit{Graph Prompt Fine-tuning}}\\
GPPT~\cite{sun2022gppt} & 48.24 ± 5.27 & 47.51 ± 7.54 & 52.86 ± 6.98 & 63.13 ± 6.78 & 52.11 ± 5.16 \\
GraphPrompt~\cite{liu2023graphprompt} & 52.48 ± 7.67 & 52.01 ± 5.54 & 56.85 ± 5.65 & 68.19 ± 7.09 & 60.05 ± 7.30 \\
GraphPrompt+~\cite{yu2024generalized} & 52.16 ± 7.84 & 52.76 ± 7.09 & 55.82 ± 6.16 & 66.31 ± 5.90 & 61.07 ± 5.37 \\
GPF~\cite{fang2023universal} & 56.16 ± 5.27 & 53.97 ± 7.45 & 55.88 ± 5.77 & 67.65 ± 6.58 & 55.69 ± 6.05 \\
ProNoG~\cite{yu2025non} & 58.06 ± 5.00 & 55.53 ± 7.58 & 56.53 ± 7.29 & 74.23 ± 5.90 & 63.54 ± 6.12 \\
\midrule
\rowcolor{gray!10}
\multicolumn{6}{c}{\textit{Multi-Domain Graph Pre-training}}\\
GCOPE~\cite{zhao2024all}  & 52.87 ± 7.15 & 55.93 ± 6.32 & 53.05 ± 7.86 & 68.80 ± 6.63 & 60.46 ± 5.56 \\
MDGPT~\cite{yu2024text} & 59.51 ± 4.54 & 55.28 ± 4.47 & 56.25 ± 6.76 & 75.73 ± 4.54 & 64.38 ± 4.13 \\
BRIDGE~\cite{yuan2025much} & \underline{63.85 ± 5.87} & \underline{60.20 ± 7.72} & 62.35 ± 5.92 & \underline{81.95 ± 5.38} & \underline{71.14 ± 4.86} \\
NodePFN~\cite{choi2026nodepfn} &  63.43 ± 7.84 &  51.02 ± 5.91 &  \underline{64.90 ± 6.90} & 76.17 ± 4.20 & 68.22 ± 6.53\\
\midrule
\textbf{\name~}(ours)  & \textbf{88.65 ± 5.54}	&\textbf{66.41 ± 6.65}	&\textbf{77.05 ± 6.80}	&\textbf{87.64 ± 3.32}	&\textbf{82.94 ± 4.78} \\
\bottomrule
\end{tabular}}
\end{center}
\vspace{-0.5cm}
\end{table*}

We conduct extensive experiments to evaluate \name~with the following research questions:
\begin{itemize}[leftmargin=*,nosep]
    \item \textbf{RQ1}: How does \name~perform on benchmark classification tasks for cross-domain GFMs?
    \item \textbf{RQ2}: How do different stages contribute to the overall performance of \name?
    \item \textbf{RQ3}: How do the hyperparameters influence the results?
\end{itemize}

\subsection{Experimental Settings}
\textbf{Dataset}. Following previous GFMs~\cite{yuan2025much}, we evaluate \name~on five benchmark graph datasets from two different domains. Specifically, 
\begin{itemize}[leftmargin=*,nosep]
    \item \textbf{Academic Domain}: includes three citation networks Cora~\cite{mccallum2000automating}, CiteSeer~\cite{giles1998citeseer}, and PubMed~\cite{sen2008collective}.
    \item \textbf{E-Commerce Domain}: consists of two Amazon product networks, Photo and Computers~\cite{shchur2018pitfalls}.
\end{itemize}

\noindent\textbf{Baselines}. We compare \name~with 16 state-of-the-art baselines from four categories, covering GFMs and related research. Specifically, 
\begin{itemize}[leftmargin=*, nosep]
    \item \textbf{End-to-end GNN}: GCN~\cite{DBLP:conf/iclr/KipfW17} and GAT~\cite{velivckovic2018graph}.
    \item \textbf{Graph Pre-training}: GCC~\cite{qiu2020gcc}, DGI~\cite{velivckovicdeep}, InfoGraph~\cite{suninfograph}, GraphCL~\cite{you2020graph}, DSSL~\cite{xiao2022decoupled} and GraphACL~\cite{xiao2023simple}.
    \item \textbf{Graph Prompt Fine-tuning}: GPPT~\cite{sun2022gppt}, GraphPrompt~\cite{liu2023graphprompt}, GraphPrompt+~\cite{yu2024generalized}, GPF~\cite{fang2023universal}, ProNoG~\cite{yu2025non}
    \item \textbf{Multi-Domain Graph Pre-training}: GCOPE~\cite{zhao2024all}, MDGPT~\cite{yu2024text}, BRIDGE~\cite{yuan2025much},  NodePFN~\cite{choi2026nodepfn}
\end{itemize}

\noindent\textbf{Multi-Domain Pre-training and Few-shot Downstream Settings}. Following previous work~\cite{yuan2025much, yu2025samgpt}, we treat each of the five datasets as an independent domain. Specifically, we use one of them as the target domain and the remaining four as source domains for pre-training. On each target domain, we perform $m$-shot node classification and graph classification, where for each class we randomly select $m$ labeled nodes or graphs as the context. For graph classification, we generate a series of graphs by constructing ego-networks centered at each labeled node in the dataset, and perform graph classification on these ego-networks, which is labeled according to its center node~\cite{yuan2025much}.

\noindent\textbf{Implementation Details}.
The graph encoder is implemented as a 3-layer GCN~\cite{DBLP:conf/iclr/KipfW17} with 256-dimensional hidden representations. For TFM, we adopt LimiX-16M architecture~\cite{zhang2025limix}. During the first stage, the unified embedding dimensionality $\tilde{d}$ is set to 50, and the graph encoder is trained for 10,000 epochs with a learning rate of 1e-3. To prevent overfitting, we employ an early stopping strategy with a patience of 50 epochs. Other implementation details are provided in Appendix.

Additional results, including using different TFM backbones, one-shot node and graph classification, other results in different few-shot setting are provided in the Appendix.

\begin{table*}[t]
\caption{The results (Accuracy ± standard deviation for one hundred runs, $\%$) of five-shot graph classification. The best results are shown in \textbf{bold} and the runner-ups are \underline{underlined}.}
\label{tab:five_shot_graph_cls}
\vspace{-0.3cm}
\begin{center}
\resizebox{\textwidth}{!}{
\begin{tabular}{lccccc}
\toprule
 Model / Target  & Cora & CiteSeer & PubMed & Photo & Computers \\
\midrule
\rowcolor{gray!15}
\multicolumn{6}{c}{\textit{End-to-end GNN}} \\ 
 GCN~\cite{DBLP:conf/iclr/KipfW17} & 51.93 ± 4.55 & 44.15 ± 5.81 & 56.88 ± 5.54 & 59.29 ± 4.78 & 51.67 ± 5.38 \\
 GAT~\cite{velivckovic2018graph} & 48.95 ± 6.08 & 43.65 ± 5.66 & 55.89 ± 5.86 & 57.18 ± 4.92 & 50.91 ± 6.22 \\
\midrule
\rowcolor{gray!15}
\multicolumn{6}{c}{\textit{Graph Self-Supervised Pre-training}} \\
 InfoGraph~\cite{suninfograph} & 52.81 ± 4.84 & 45.51 ± 5.44 & 60.47 ± 6.25 & 62.18 ± 5.24 & 53.40 ± 5.49 \\
 GraphCL~\cite{you2020graph} & 53.92 ± 4.65 & 48.47 ± 4.91 & 61.84 ± 6.48 & 62.74 ± 5.19 & 57.73 ± 6.27 \\
 DSSL~\cite{xiao2022decoupled} & 55.03 ± 5.57 & 45.58 ± 6.20 & 62.45 ± 4.95 & 63.04 ± 6.13 & 57.26 ± 6.33 \\
 GraphACL~\cite{xiao2023simple} & 55.52 ± 5.83 & 47.26 ± 5.24 & 61.44 ± 5.90 & 63.69 ± 6.33 & 56.54 ± 6.46 \\
\midrule
\rowcolor{gray!15}
\multicolumn{6}{c}{\textit{Graph Prompt Fine-tuning}} \\
GraphPrompt~\cite{liu2023graphprompt} & 55.24 ± 6.39 & 52.31 ± 5.04 & 63.09 ± 6.22 & 64.27 ± 5.86 & 58.12 ± 5.19 \\
GraphPrompt+~\cite{yu2024generalized} & 59.53 ± 5.07 & 51.22 ± 5.88 & 62.81 ± 5.83 & 65.73 ± 4.60 & 58.61 ± 6.24 \\
GPF~\cite{fang2023universal} & 56.19 ± 5.28 & 53.47 ± 5.33 & 61.39 ± 5.93 & 64.08 ± 5.24 & 57.47 ± 5.96 \\
ProNoG~\cite{yu2025non} & 61.07 ± 5.31 & 51.25 ± 5.98 & 64.73 ± 6.12 & 66.10 ± 5.14 & 60.55 ± 5.17 \\
\midrule
\rowcolor{gray!15}
\multicolumn{6}{c}{\textit{Multi-Domain Graph Pre-training}} \\
GCOPE~\cite{zhao2024all} & 57.29 ± 4.77 & 53.97 ± 6.13 & 62.61 ± 4.94 & 65.77 ± 5.35 & 58.20 ± 4.65 \\
MDGPT~\cite{yu2024text} & 61.77 ± 4.10 & 55.37 ± 6.33 & 62.70 ± 6.20 & 67.06 ± 3.72 & 59.76 ± 8.56 \\
BRIDGE~\cite{yuan2025much} & \underline{67.21 ± 3.78} & \underline{59.80 ± 5.04} & \underline{67.17 ± 3.77} & \underline{70.53 ± 6.53} & \underline{64.91 ± 6.24} \\
\midrule
\textbf{\name~}(ours) & \textbf{83.95 ± 5.39}	& \textbf{64.49 ± 7.21}	& \textbf{73.15 ± 7.08}	& \textbf{81.79 ± 4.66}	& \textbf{72.97 ± 6.30}\\
\bottomrule
\end{tabular}}
\end{center}
\vspace{-0.2cm}
\end{table*}

\subsection{Comparisons with State-of-the-art}
To answer \textbf{RQ1}, we report results of comparing \name~with baselines on both node and graph classification in few-shot setting. More experiments results are in Appendix.

\noindent\textbf{Five-shot Node Classification}. As shown in Table~\ref{tab:five_shot_node_cls}, \name~consistently and significantly outperforms all baselines across all evaluated datasets. On the Cora dataset, our model achieves an accuracy of 88.65\%, representing a 24.80\% absolute improvement over the strongest baseline BRIDGE. Besides, on the node-rich PubMed and Computers datasets, our method maintains superior accuracy. Compared with NodePFN, which learns posterior predictive distributions from synthetic graph priors, \name~achieves better performance across all target dataset. This suggests that synthetic-prior-based graph PFN pretraining alone may not be sufficient for cross-domain few-shot setting on real datasets. In contrast, \name~introduces community-guided continual pre-training to expose the TFM to topology-derived few-shot prediction episodes, explicitly aligning graph topology with the tabular representation space and thereby enabling more effective prediction.

\noindent\textbf{Five-shot Graph Classification}. As illustrated in Table~\ref{tab:five_shot_graph_cls}, \name~comprehensively outperforms all baselines in graph classification tasks, with accuracy exceeding 80\% on both Cora and Photo. These results demonstrate that our approach yields more trustworthy predictions for graph-level tasks than current state-of-the-art multi-domain pre-training methods, while maintaining standard deviations on par with other method BRIDGE. When considered alongside the node-level experiments, these findings provide strong evidence that \name~exhibits remarkable few-shot stability across downstream tasks.

\begin{figure*}[t]
  \centering
  \begin{subfigure}[b]{0.23\textwidth}
    \centering
    \includegraphics[width=\linewidth]{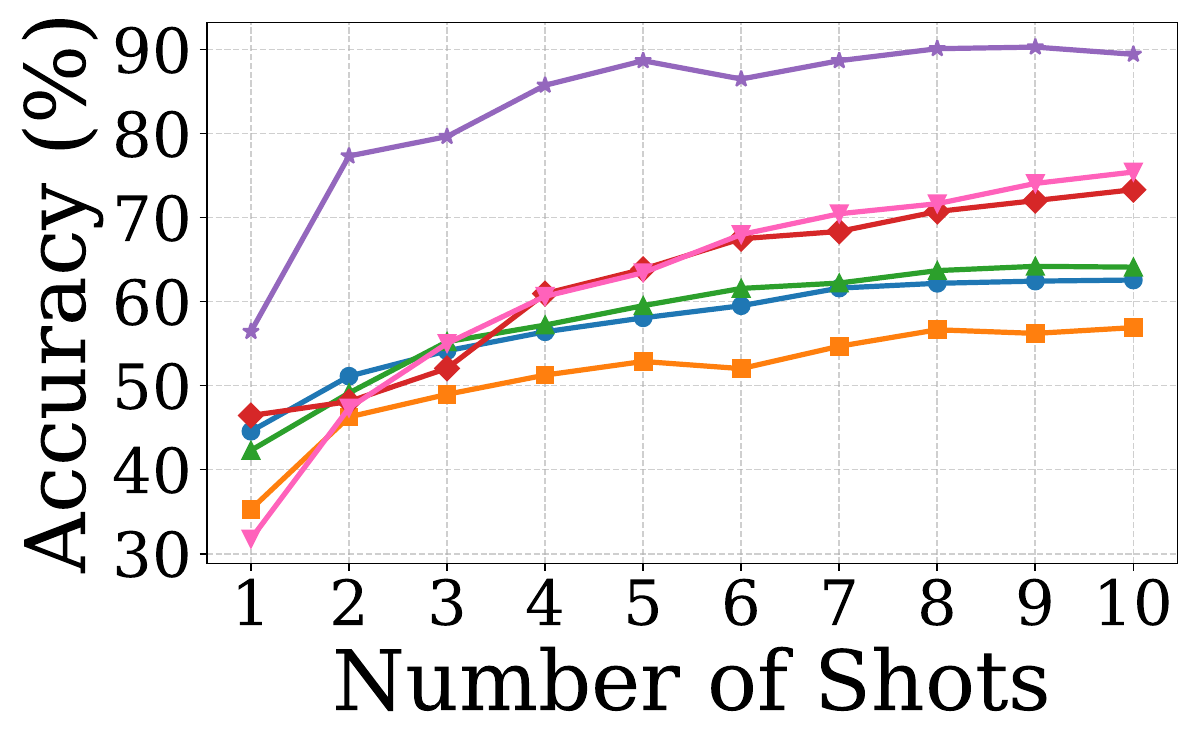}
    \caption{Cora dataset}
    \label{fig:cora_node}
  \end{subfigure}
  \hfill
  \begin{subfigure}[b]{0.23\textwidth}
    \centering
    \includegraphics[width=\linewidth]{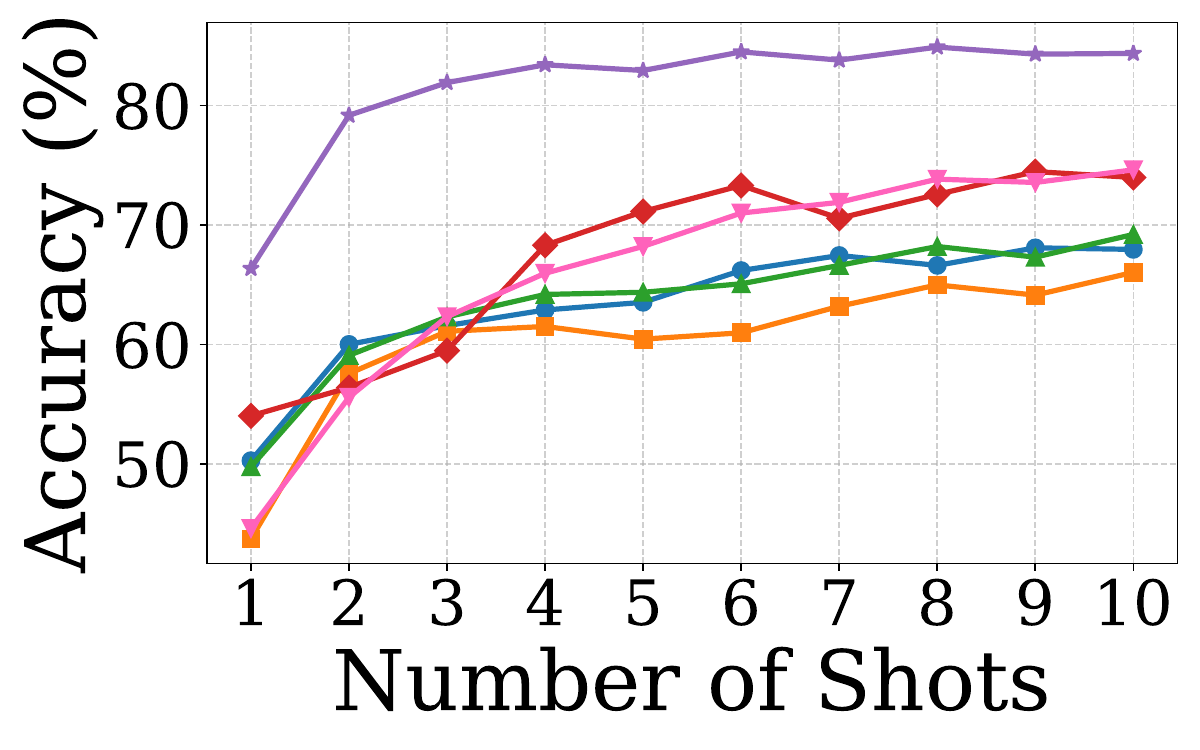}
    \caption{Computers dataset}
    \label{fig:compters_node}
  \end{subfigure}
  \hfill
  \begin{subfigure}[b]{0.23\textwidth}
    \centering
    \includegraphics[width=\linewidth]{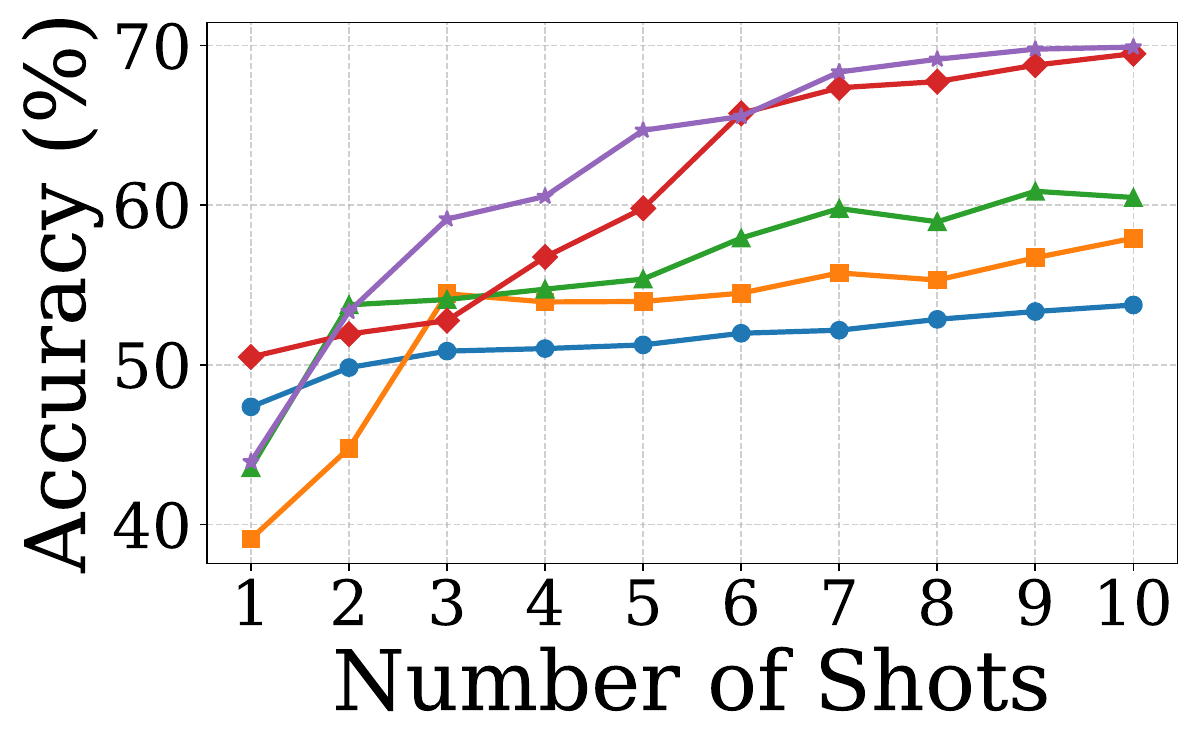}
    \caption{CiteSeer dataset}
    \label{fig:citeseer_graph}
  \end{subfigure}
  \hfill
  \begin{subfigure}[b]{0.23\textwidth}
    \centering
    \includegraphics[width=\linewidth]{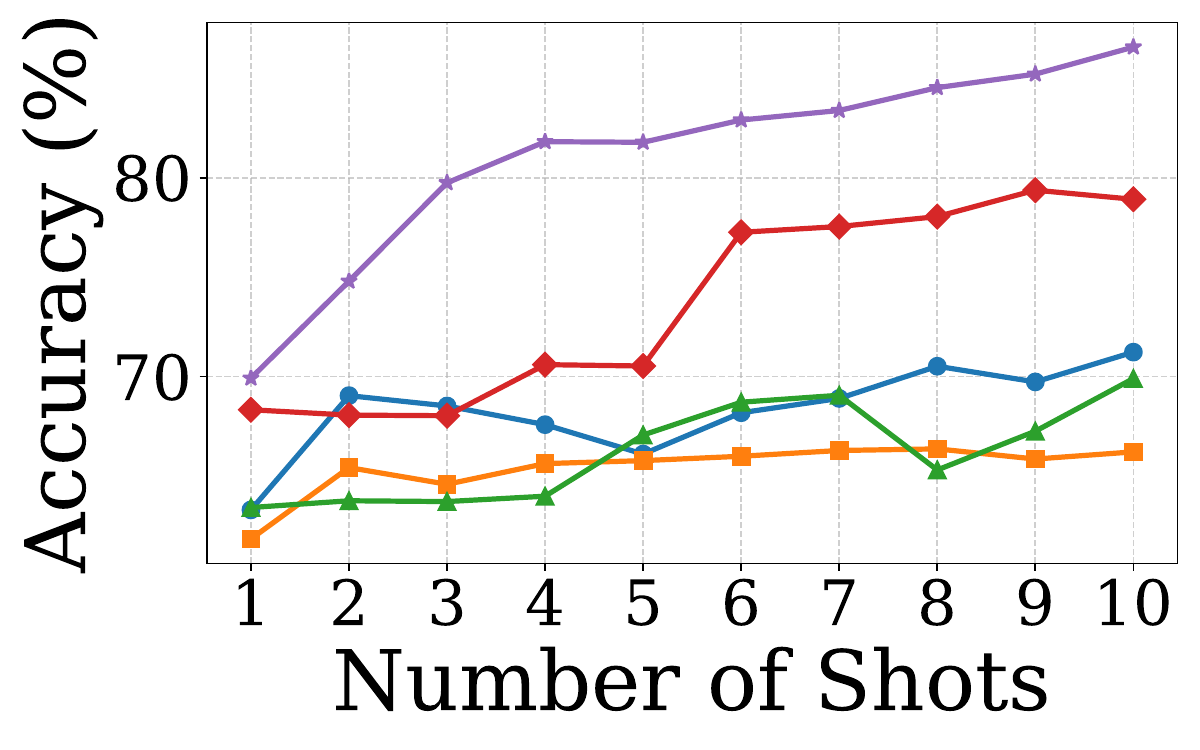}
    \caption{Photo dataset}
    \label{fig:photo_graph}
  \end{subfigure}

  \medskip
  \begin{minipage}{0.98\textwidth}
    \centering
    \includegraphics[width=\linewidth]{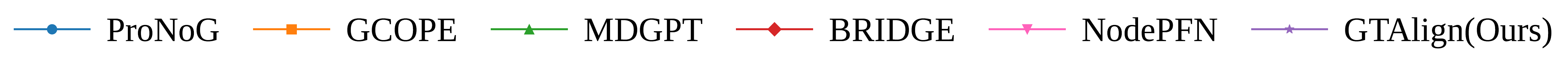}
  \end{minipage}
  \vspace{-0.15cm}
  \caption{$m$-shot performance for node and graph classification, corresponding to (a)-(b) and (c)-(d), respectively.}
  \label{fig:mshot_exp}
  \vspace{-0.15cm}
\end{figure*}

\noindent\textbf{Different Few-Shot Scenarios}.
To further investigate the capabilities of \name~in few-shot scenarios, we further evaluate the performance under different few-shot settings, where $m$ ranges from 1 to 10, as shown in Figure~\ref{fig:cora_node} and Figure~\ref{fig:compters_node} for node classification, and Figure~\ref{fig:citeseer_graph} and Figure~\ref{fig:photo_graph} for graph classification. For node classification, taking Cora and Computers as examples, \name~consistently outperforms other GFM methods in few-shot scenarios with varying amounts of labeled data. For graph classification, when CiteSeer is the target domain, our method falls behind most baselines in the 1-shot scenario when the target dataset in the dataset. This discrepancy is intrinsically attributed to the nature of graph-level prediction, which typically relies on aggregating node-level information to form a holistic graph representation. When contextual examples are extremely limited, the aggregated node information is statistically insufficient, compelling the model to make conservative predictions that lean toward the pre-trained prior. As the number of context samples increases, the aggregated graph representations become progressively more stable, enabling the TFM to effectively leverage its pre-trained prior and rapidly converge to accurate predictions.

\begin{figure}[t]  
  \centering
  \begin{minipage}{0.48\columnwidth}
    \centering
    \includegraphics[width=0.95\linewidth]{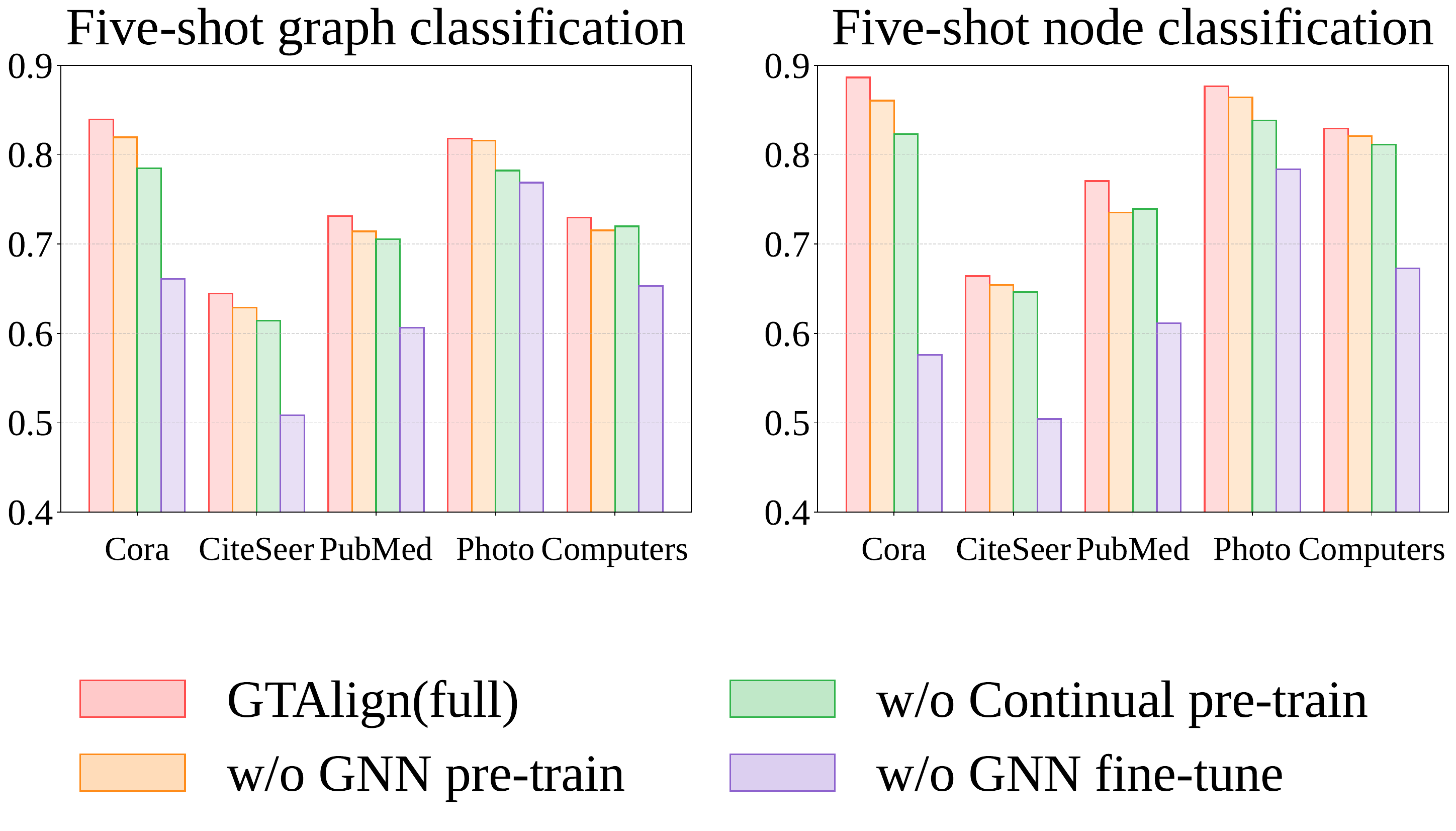}
    \captionof{figure}{Ablation studies. Five-shot results on graph (left) and node (right) classification.}
    \label{fig:ablation_fig}
  \end{minipage}%
  \hfill
  \begin{minipage}{0.49\columnwidth}
    \centering
    \refstepcounter{figure}%
    \begin{subfigure}[b]{0.48\linewidth}
      \centering
      \includegraphics[width=\linewidth]{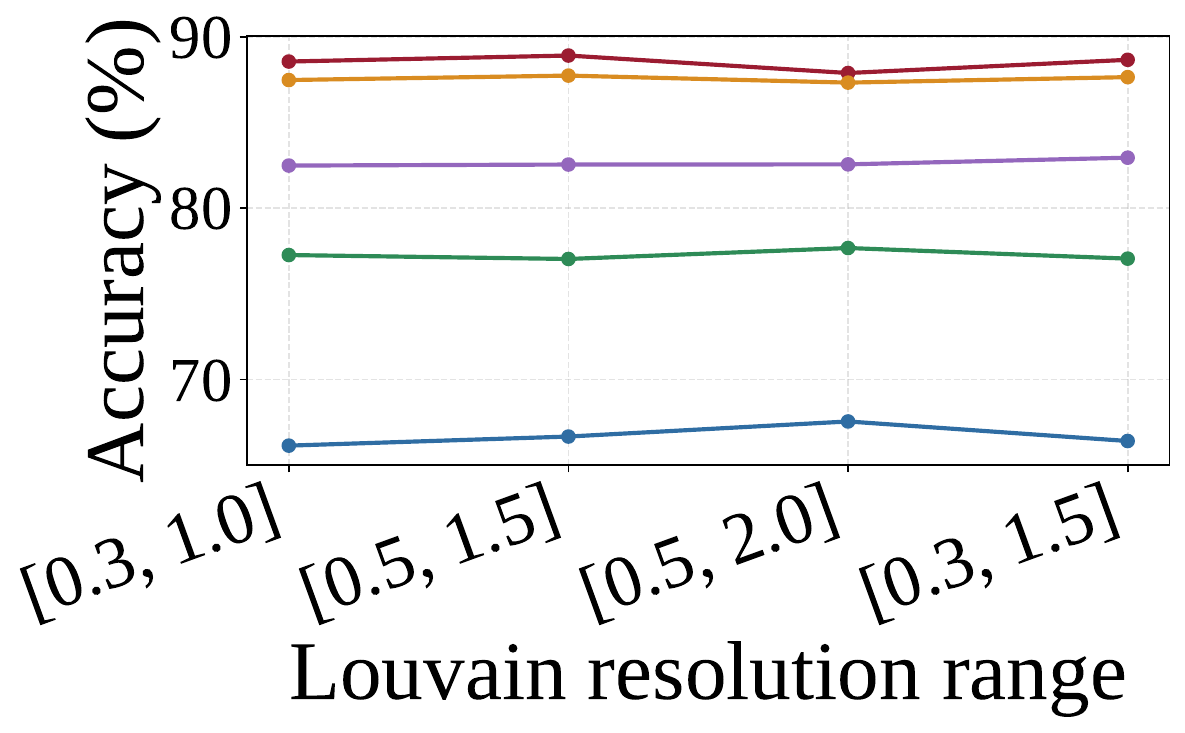}
      \caption{Louvain resolution}
      \label{fig:sensitivity_resolution}
    \end{subfigure}%
    \hfill
    \begin{subfigure}[b]{0.48\linewidth}
      \centering
      \includegraphics[width=\linewidth]{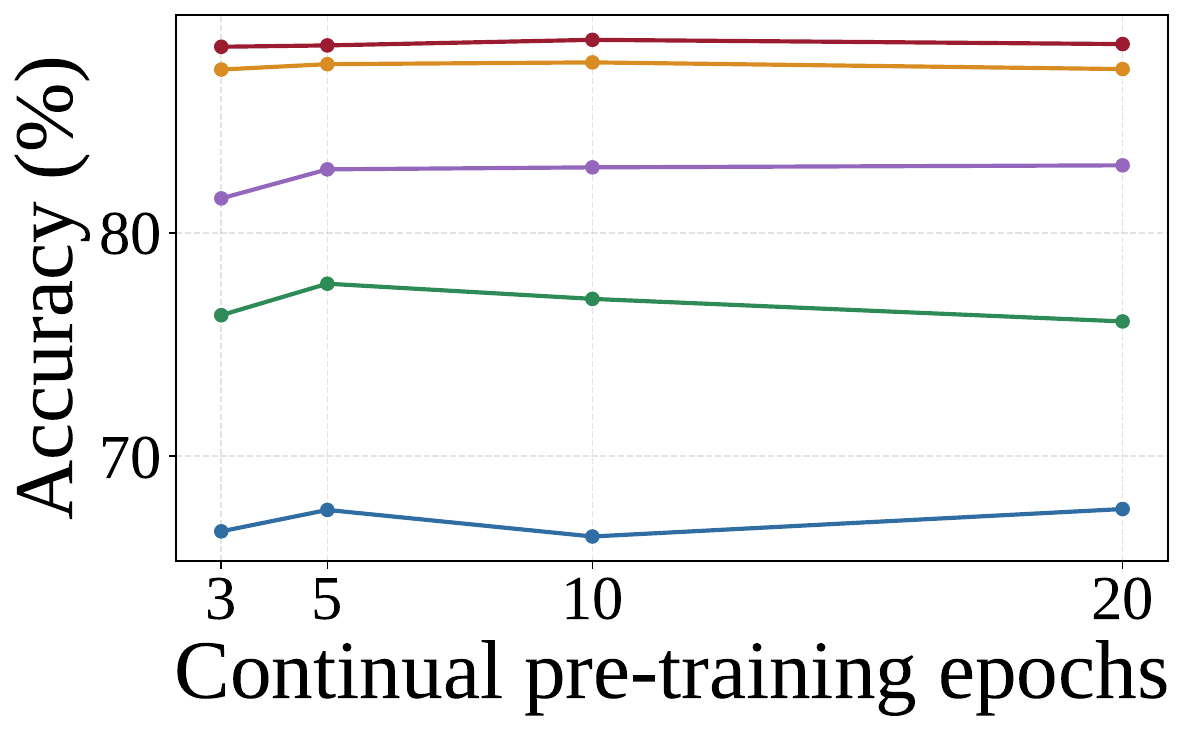}
      \caption{Epoch range}
      \label{fig:sensitivity_epoch}
    \end{subfigure}

    \medskip
    \includegraphics[width=\linewidth]{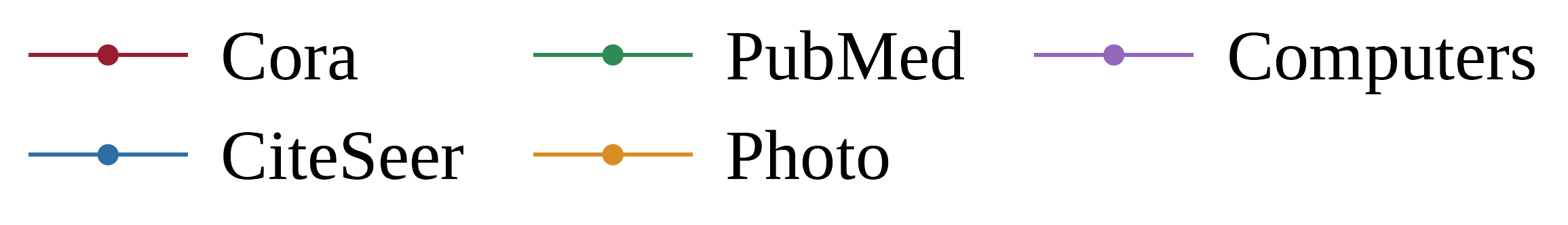}
    \vspace{-0.5cm}
    \addtocounter{figure}{-1}%
    \captionof{figure}{Hyperparameter sensitivity analysis of SET-GFM under the five-shot setting.}
    \label{fig:hyperparameter_sensitivity}

  \end{minipage}
\vspace{-0.5cm}
\end{figure}

\subsection{Ablation Studies}
To answer \textbf{RQ2}, we conduct ablation studies across three key components of our framework. 
\begin{itemize}[nosep, leftmargin=*] 
    \item w/o GNN pre-train: We use a randomly initialized GNN encoder to replace the pre-trained one.
    \item w/o Continual pre-train: We remove the community-guided continual pre-training stage.
    \item w/o GNN fine-tune: We use the pre-trained encoder weights without target-domain fine-tuning.
\end{itemize}

The experimental results under five-shot for both node and graph classification tasks are summarized in Figure~\ref{fig:ablation_fig}. Overall, removing GNN fine-tuning stage on the target-domain leads to the most significant performance degradation, especially on node classification, indicating that adapting the graph encoder with a few labeled target samples is crucial for producing task-relevant representations for TFM-based inference. Removing the continual pre-training stage also consistently reduces performance, demonstrating the effectiveness of Louvain-based pseudo-labeling of the second stage. In contrast, removing the initial GNN pre-training causes relatively mild degradation, suggesting that while multi-domain graph pre-training provides a useful initialization, the subsequent graph-to-table alignment and target-domain adaptation play more dominant roles in the final performance. We have also provided specific numerical tables in Appendix.

\subsection{Hyperparamter Sensitivity Analysis}

To answer \textbf{RQ3}, we further evaluate the sensitivity of SET-GFM to the Louvain resolution range while fixing the continual pre-training epoch to 10. As shown in Figure~\ref{fig:sensitivity_resolution}, SET-GFM achieves stable performance across different resolution ranges. This robustness suggests that exposing the model to communities at different granularities is beneficial, while the method does not require careful tuning of the Louvain resolution parameter. We also analyze the sensitivity of SET-GFM to the number of continual pre-training epochs. As shown in Figure~\ref{fig:sensitivity_epoch}, the performance remains stable across different epoch settings. Increasing the number of continual pre-training epochs generally improves performance on Cora and CiteSeer, while the results on PubMed and Photo remain relatively stable. This indicates that the proposed continual pre-training stage is robust to the choice of training epochs.

\section{Related Work}

\textbf{Graph Foundation Model.}
GFMs are conceived as a model pre-trained on massive graph data that can adapt to a variety of downstream graph tasks~\cite{liu2025graph}. Based on its backbone architecture, the existing methods can be mainly divided into three groups: GNN-based, LLM-based, and GNN+LLM-based. Among the GNN-based methods, GraphPrompt~\cite{liu2023graphprompt} unifies pre-training and downstream tasks into a consistent subgraph similarity format, employing task-specific prompt vectors to narrow the objective gap.  Building on this, GraphPrompt+~\cite{yu2024generalized} enhances knowledge extraction by generalizing pre-training tasks and adopting hierarchical prompts to capture multi-level structural information. GCOPE~\cite{zhao2024all} further extends this paradigm to few-shot learning across diverse datasets. OMOG~\cite{liu2024one} tackles cross‑graph transfer by pretraining a separate source model for each pretraining graph and storing them in a model bank. SAMGPT~\cite{yu2025samgpt} introduces structure tokens to align structural distributions across source domains during pre‑training, and employs holistic and specific prompts to adapt both shared and domain‑specific structural knowledge to an unseen target domain. BRIDGE~\cite{yuan2025much} unifies multi-domain graph features through a domain-invariant feature aligner and utilizes a lightweight mixture-of-experts network to maintain generalization across diverse tasks. However, a common limitation of these GNN-based methods is that they often lack pre-training on truly massive and diverse datasets required to function as a universal foundation model.

For LLM-based methods, researchers attempt to convert the structural information of graphs into natural language descriptions, directly using them as input to the LLM. 
Representative works include NLGraph~\cite{wang2023can}, GPT4Graph~\cite{guo2023gpt4graph}, and InstructGLM~\cite{ye2024language}. In GNN+LLM-based methods, researchers employ the advantage of GNNs in capturing topological structures and LLMs in semantic reasoning, such as OFA~\cite{liuone}, UniGraph~\cite{he2025unigraph}, and GFT~\cite{wang2024gft}. However, these methods are fundamentally predicated on the availability of textual information, facing severe challenges when applied to graphs without sufficient textual information. Therefore, how to design graph-based models that do not rely on explicit textual attributes remains a key challenge for GFMs.

\textbf{Tabular Foundation Model.} Recently, the emergence of PFNs, notably TabPFN~\cite{hollmann2025accurate} and LimiX~\cite{zhang2025limix}, has established a new paradigm for TFMs, demonstrating superior performance through in-context inference on small-to-medium-scale tabular data. Following this success, several studies have extended the PFN framework to diverse domains, such as time-series forecasting~\cite{hoo2025tables} and causal learning~\cite{balazadeh2025causalpfn}. In addition, following PFN, some works attempt to train GFMs from synthetic data~\cite{eremeev2025graphpfn, choilearning}. For graph data, pioneering studies have successfully extended TFMs to specialized graph tasks, including node classification~\cite{eremeevturning, eremeev2025graphpfn, hayler2025graphs, choi2025can}, graph anomaly detection~\cite{liu2026tabular}, and link prediction~\cite{liao2026tfmlinker}. However, existing methods mostly focus on single-domain tasks by concatenating graph topology-based features as inputs to TFM, which limits their ability to handle multi-domain GFMs. We tackle this challenge by combining graph structure encoders with TFM and design a self-supervised continuous pre-training to bridge the gap between non-Euclidean topology and tabular-based models.

\section{Conclusion}
In this paper, we introduce \name, a simple yet effective graph-to-table alignment framework for text-free graph foundation models. First, we pre‑train a graph encoder that learn unified graph representations in a latent space. Then, we align graph structural with the tabular representation space via community-guided continual pre-training. Finally, \name~adapts the graph encoder for the target domain and performs in-context inference. Experiments on five datasets show that \name~consistently outperforms state-of-the-art baselines on both node and graph classification.

In the future, we plan to test our method for more graph learning tasks~\cite{stoll2025graphbench,gastinger2024tgb} and more application scenarios such as relational deep learning~\cite{chen2025relgnn}.

\bibliography{main}
\bibliographystyle{iclr2026_conference}

\appendix
\newpage
\section{Appendix}

\subsection{Experimental Settings}

\begin{table*}[h]
\caption{Datasets statistics.}
\label{tab:datasets_statistics}
\centering
\resizebox{\linewidth}{!}{%
\begin{tabular}{lcccccll}
\toprule
Dataset & \#Nodes & \#Edges & \#Features & \#Classes & \#Full-shot & Domain & Source \\
\midrule
Cora & 2,708 & 10,556 & 1,433 & 7 & 140 & Academic  & ~\cite{mccallum2000automating} \\
CiteSeer & 3,327 & 9,104 & 3,703 & 6 & 120 & Academic  & ~\cite{giles1998citeseer}  \\
PubMed & 19,717 & 88,651 & 500 & 3 & 60 & Academic &~\cite{sen2008collective}\\
Amazon-Photo & 7,650 & 238,162 & 745 & 8 & 160 & E-Commerce & ~\cite{shchur2018pitfalls} \\
Amazon-Computers & 13,752 & 491,722 & 767 & 10 & 200 & E-Commerce & ~\cite{shchur2018pitfalls} \\
\bottomrule
\end{tabular}}

\end{table*}

\noindent\textbf{Datasets}. We utilise 5 publicly available graph benchmark datasets from different domains, as shown in Table~\ref{tab:datasets_statistics}, including:
\begin{itemize}[leftmargin=0.5cm]
    \item Cora~\cite{mccallum2000automating} is a citation network where nodes represent scientific publications and edges denote citation relationships. Publications are classified into 7 research fields.
    \item CiteSeer~\cite{giles1998citeseer} is also a citation network of scientific publications. Nodes are papers connected by citation links, and papers belong to 6 categories.
    \item PubMed~\cite{sen2008collective} is a citation network of biomedical publications. Nodes are papers, edges are citations, and papers are divided into 3 classes.
    \item Amazon-Photo and Amazon-Computers~\cite{shchur2018pitfalls} are co-purchase networks. Nodes represent products, edges indicate that two products are frequently bought together. The two datasets contain 8 and 10 product categories, respectively.
\end{itemize}

Following existing work~\cite{yu2025samgpt}, we fixed the random seed at 42 during the data partitioning phase and used stratified sampling by category to construct a few-shot training set. Specifically, for each category, k samples were randomly selected without replacement in each sampling as a support/context set, and this sampling was repeated multiple times to obtain different partitions, thereby reducing the randomness of a single partition and improving the comparability and reproducibility of the results. We have also open-sourced this part.

\noindent\textbf{Implementation Details}.
The graph encoder is implemented as a 3-layer GCN~\cite{DBLP:conf/iclr/KipfW17} with 256-dimensional hidden representations. For TFM, we adopt LimiX-16M architecture~\cite{zhang2025limix}. During the first stage, the unified embedding dimensionality $\tilde{d}$ is set to 50, and the graph encoder is trained for 10,000 epochs with a learning rate of 1e-3. To prevent overfitting, we employ an early stopping strategy with a patience of 50 epochs. In the second stage, we jointly optimize all trainable parameters with Adam at learning rate 1e-4 for 10 epochs. In each epoch, every source graph contributes 5 random episodes. For each episode, we run Louvain on the raw adjacency with resolution $\rho \sim \mathrm{Uniform}(0.3, 1.5)$. We save the checkpoint that achieves the highest average pseudo-task accuracy across episodes. In the third stage, we optimize the parameters for 100 epochs using a learning rate of 1e-3, and employ an early stopping strategy. For node classification and graph classification respectively, we generate 100 $m$-shot tasks by repeatedly sampling $m$ labeled nodes or graphs per class for 100 times. Accuracy is used as the evaluation metric for each task, and the mean accuracy and standard deviation of these 100 results are reported. All models are implemented using PyTorch 2.9.0 with CUDA 12.8. The experiments are executed on NVIDIA 5090 GPUs.

\noindent\textbf{Baselines}. 
We compare our method with 16 state-of-the-art baselines from four categories, covering GFMs and related research, including End-to-end GNN, Graph Self-Supervised Pre-training, Graph Prompt Fine-tuning and Multi-Domain Graph Pre-training.

\noindent\textit{End-to-end GNN}. This category includes graph neural networks trained directly on the downstream target domain without pre-training, and they are applied in both node and graph classifications.
\begin{itemize}[leftmargin=0.5cm]
    \item GCN~\cite{DBLP:conf/iclr/KipfW17} is a spectral graph neural network that performs neighborhood aggregation through layer-wise graph convolution. It captures local graph structure by propagating and transforming node features along graph edges.
    \item GAT~\cite{velivckovic2018graph} introduces attention mechanisms into neighborhood aggregation, allowing the model to assign different importance weights to neighboring nodes.
\end{itemize}

\noindent\textit{Graph Self-Supervised Pre-training}. This category includes methods that first pre-train graph encoders with self-supervised objectives and then adapt them to downstream tasks.
\begin{itemize}[leftmargin=0.5cm]
    \item GCC~\cite{qiu2020gcc} learns transferable structural representations through contrastive learning over subgraphs sampled from multiple networks. It is used for node classifications.
    \item DGI~\cite{velivckovicdeep} maximizes mutual information between local node representations and a global graph summary, encouraging the encoder to capture informative node-level features. It is applied in node classification.
    \item InfoGraph~\cite{suninfograph} learns discriminative whole-graph embeddings by maximizing the mutual information between global graph representations and representations of local substructures (e.g., nodes, edges) at different scales. It is applied to graph classification.
    \item GraphCL~\cite{you2020graph} performs graph contrastive learning by maximizing agreement between different augmented views of the same graph. It is applied in both node and graph classification.
    \item DSSL~\cite{xiao2022decoupled} is a generative self-supervised learning framework that does not rely on the graph homophily assumption. By decoupling node latent semantics from the graph generation process, it effectively addresses the failure of traditional graph self-supervised learning methods on heterophilous graphs. It is applied to both node and graph classification.
    \item GraphACL~\cite{xiao2023simple} requires neither graph augmentations nor homophily assumptions, which captures both one-hop neighborhood context and two-hop monophily by making each node predict the representations of its one-hop neighbors, achieving superior performance on both homophilic and heterophilic graphs. It is applied to both node and graph classification.
\end{itemize}

\noindent\textit{Graph Prompt Fine-tuning}. In graph prompt fine-tuning, prompt-based methods leverage learnable, task-specific prompts to steer the fine-tuning of pre-trained models, thereby enhancing their transferability and adaptability to downstream tasks.
\begin{itemize}[leftmargin=0.5cm]
    \item GPPT~\cite{sun2022gppt} uses masked edge prediction as a pre-training task and reformulates node classification into link prediction via a graph prompt function, thereby bridging the objective gap between pre-training and downstream tasks and significantly improving generalization performance in few-shot scenarios. It is applied to both node classification.
    \item GraphPrompt~\cite{liu2023graphprompt} unifies pre-training and downstream tasks via learnable graph prompts, enabling pre-trained models to directly adapt to various downstream graph tasks without fine-tuning. It is applied to both node and graph classification.
    \item GraphPrompt+~\cite{yu2024generalized} unifies pre-training and downstream tasks on graphs by transforming task-specific inputs into a consistent prompting format, eliminating the need for fine-tuning. It is applied to both node and graph classification.
    \item GPF~\cite{fang2023universal} operates directly on the input graph's feature space, enabling effective adaptation of pre-trained GNN models under any pre-training strategy without the need for task-specific prompt designs. It is applied to both node and graph classifications.
    \item ProNoG~\cite{yu2025non} is a pre-training and prompt learning framework for non-homophilic graphs, which leverages a conditional network to generate node-specific prompts, effectively capturing fine-grained non-homophilic patterns and achieving superior few-shot node and graph classification performance. It is applied to both node and graph classification.
\end{itemize}

\noindent\textit{Multi-Domain Graph Pre-training}. This line of work aims to learn graph representations that transfer well across different domains, making these methods the most comparable baselines to our approach. 
\begin{itemize}[leftmargin=0.5cm]
    \item GCOPE~\cite{zhao2024all} is a cross-domain graph pretraining framework that introduces learnable coordinator virtual nodes for each graph dataset and establishes inter-graph connections, unifying heterogeneous graphs into a single communication network to mitigate negative transfer and enable effective multi-domain knowledge integration and transfer. It is applied to both node and graph classification.
    \item MDGPT~\cite{yu2024text} is a text-free multi-domain graph pre-training framework that uses domain tokens to unify features and dual prompts to adapt to target domains, mitigating domain conflicts and improving node and graph classification performance. It is applied to both node and graph classification.
    \item BRIDGE~\cite{yuan2025much} is a multi-domain graph pre-training and prompt learning framework that aligns features with domain-invariant aligners, adaptively transfers knowledge via a lightweight mixture-of-experts network. It is applied to both node and graph classifications.
    \item NodePFN~\cite{choi2026nodepfn} introduces a universal node classification model, by training exclusively on synthetic graphs generated from carefully designed priors, learns to perform in-context inference on arbitrary real-world graphs without any dataset-specific fine-tuning. It is applied to both node classification.
\end{itemize}

\begin{table*}[ht]
\caption{Accuracy ($\% ±$ standard deviation for one hundred runs) of one-shot node classification. The best results are shown in \textbf{bold} and the runner-ups are \underline{underlined}.}
\label{tab:one_shot_node_cls}
\centering
\resizebox{\textwidth}{!}{%
\begin{tabular}{lcccccc}
\toprule
Model / Target  & Cora & CiteSeer & PubMed & Photo & Computers &Rank\\
\midrule
\rowcolor{gray!15}
\multicolumn{7}{c}{\textit{End-to-end GNN}}
\\
GCN~\cite{DBLP:conf/iclr/KipfW17} & 29.36 ± 4.20 & 30.99 ± 4.85 & 40.94 ± 7.05 & 40.05 ± 7.25 & 34.62 ± 8.95 & 15.2  \\
GAT~\cite{velivckovic2018graph} & 29.00 ± 5.33 & 29.46 ± 3.32 & 40.09 ± 5.21 & 35.60 ± 6.52 & 33.15 ± 6.59 & 16.4\\
\midrule
\rowcolor{gray!15}
\multicolumn{7}{c}{\textit{Graph Self-Supervised Pre-training}}\\
GCC~\cite{qiu2020gcc} & 31.67 ± 5.23 & 32.55 ± 2.69 & 41.66 ± 4.58 & 42.10 ± 5.99 & 35.91 ± 5.68 & 12.6\\
DGI~\cite{velivckovicdeep} & 30.82 ± 4.41 & 31.85 ± 4.36 & 40.08 ± 6.22 & 47.23 ± 6.03 & 37.05 ± 6.40  & 13.2\\
GraphCL~\cite{you2020graph}  & 33.28 ± 6.03 & 29.12 ± 4.26 & 39.31 ± 7.05 & 42.98 ± 6.54 & 42.87 ± 5.32 & 13.2 \\
DSSL~\cite{xiao2022decoupled} & 30.65 ± 5.24 & 31.20 ± 6.33 & 40.89 ± 6.94 & 45.32 ± 7.28 & 38.41 ± 6.54 & 13.6\\
GraphACL~\cite{xiao2023simple} & 35.26 ± 4.65 & 34.09 ± 6.40 & 43.54 ± 5.58 & 49.20 ± 7.00 & 41.86 ± 6.20 & 9.6\\
\midrule
\rowcolor{gray!15}
\multicolumn{7}{c}{\textit{Graph Prompt Fine-tuning}}\\
GPPT~\cite{sun2022gppt} & 32.18 ± 5.02 & 31.33 ± 4.20 & 41.27 ± 5.60 & 46.98 ± 4.04 & 35.18 ± 8.09 & 12.6\\
GraphPrompt~\cite{liu2023graphprompt} & 37.95 ± 6.31 & 34.92 ± 6.75 & 45.85 ± 8.58 & 50.42 ± 8.05 & 42.58 ± 7.21 & 7.6\\
GraphPrompt+~\cite{yu2024generalized} & 36.06 ± 6.93 & 33.85 ± 7.62 & 45.08 ± 7.99 & 52.28 ± 7.51 & 43.39 ± 6.90 & 7.2\\
GPF~\cite{fang2023universal} & 40.26 ± 8.33 & 40.20 ± 7.10 & 47.33 ± 6.63 & 51.48 ± 5.34 & 40.09 ± 6.19 & 6.4\\
ProNoG~\cite{yu2025non} & 44.57 ± 6.57 & 39.96 ± 7.99 & 50.48 ± 7.06 & 63.30 ± 4.92 & 50.29 ± 6.32 & 3.8 \\
\midrule
\rowcolor{gray!15}
\multicolumn{7}{c}{\textit{Multi-Domain Graph Pre-training}}\\
GCOPE~\cite{zhao2024all} & 35.29 ± 4.29 & 40.75 ± 4.28 & 44.55 ± 6.07 & 51.55 ± 5.36 & 43.74 ± 7.05 & 6.2\\
MDGPT~\cite{yu2024text} & 42.29 ± 7.75 & 37.32 ± 7.01 & 50.89 ± 7.74 & 63.63 ± 7.23 & 49.78 ± 8.77 & 4.0\\
BRIDGE~\cite{yuan2025much} & \underline{46.44 ± 8.01} & \underline{42.18 ± 8.89} & \textbf{56.35 ± 7.22} & \underline{67.87 ± 7.45} & \underline{54.04 ± 8.52} & \underline{1.8}\\
NodePFN~\cite{choi2026nodepfn} & 31.59 ± 7.89 & 36.79 ± 7.39 & 43.93 ± 6.58 & 51.31 ± 9.80 & 44.27 ± 9.77 & 8.4 \\
\midrule
\textbf{\name~}(ours) & \textbf{56.43 ± 13.49} & \textbf{43.41 ± 10.26} & \underline{52.00 ± 11.29} & \textbf{79.52 ± 10.17} & \textbf{66.37 ± 11.48} & \textbf{1.2}\\
\bottomrule
\end{tabular}}

\end{table*}

\begin{table*}
\caption{Accuracy ($\% ±$ standard deviation for one hundred runs) of one-shot graph classification. The best results are shown in \textbf{bold} and the runner-ups are \underline{underlined}.}
\label{tab:one_shot_graph_cls}
\centering
\resizebox{\textwidth}{!}{%
\begin{tabular}{lcccccc}
\toprule
Model / Target  & Cora & CiteSeer & PubMed & Photo & Computers & Rank  \\
\midrule
\rowcolor{gray!15}
\multicolumn{7}{c}{\textit{End-to-end GNN}}\\
GCN~\cite{DBLP:conf/iclr/KipfW17} & 38.59 ± 6.33 & 29.90 ± 7.20 & 47.26 ± 7.12 & 56.02 ± 5.28 & 39.29 ± 6.25 & 12.8\\
GAT~\cite{velivckovic2018graph} & 35.64 ± 5.09 & 26.78 ± 6.91 & 41.72 ± 6.99 & 50.23 ± 4.98 & 37.25 ± 6.73  & 14.0\\
\midrule
\rowcolor{gray!15}
\multicolumn{7}{c}{\textit{Graph Self-Supervised Pre-training}}\\
InfoGraph~\cite{suninfograph} & 41.76 ± 4.89 & 31.03 ± 5.07 & 47.92 ± 6.76 & 59.88 ± 5.09 & 41.33 ± 6.67  & 10.4\\
GraphCL~\cite{you2020graph} & 40.23 ± 5.28 & 32.99 ± 5.29 & 49.21 ± 8.16 & 58.02 ± 5.11 & 42.09 ± 5.38 & 10.6\\
DSSL~\cite{xiao2022decoupled} & 40.95 ± 6.29 & 32.03 ± 6.80 & 49.32 ± 7.13 & 58.19 ± 6.10 & 41.94 ± 6.23 & 10.4\\
GraphACL~\cite{xiao2023simple} & 41.08 ± 5.77 & 32.38 ± 5.91 & 49.25 ± 6.22 & 59.66 ± 5.27 & 43.18 ± 5.99 & 9.4\\
\midrule
\rowcolor{gray!15}
\multicolumn{7}{c}{\textit{Graph Prompt Fine-tuning}}\\
GraphPrompt~\cite{liu2023graphprompt} & 42.18 ± 4.31 & 38.66 ± 6.12 & 51.28 ± 5.29 & 60.07 ± 5.70 & 47.31 ± 5.11 & 6.8\\
GraphPrompt+~\cite{yu2024generalized} & 43.87 ± 5.18 & 40.07 ± 6.21 & 51.83 ± 5.69 & 61.30 ± 5.91 & 47.08 ± 6.32 & 5.8\\
GPF~\cite{fang2023universal} & 41.82 ± 8.01 & 39.00 ± 7.45 & 47.23 ± 8.90 & 59.21 ± 5.57 & 46.03 ± 9.01 & 9.2\\
ProNoG~\cite{yu2025non} & 50.98 ± 7.33 & \underline{47.37 ± 7.08} & \underline{55.35 ± 7.90} & 63.29 ± 6.33 & 50.21 ± 7.29 & 3.0\\
\midrule
\rowcolor{gray!15}
\multicolumn{7}{c}{\textit{Multi-Domain Graph Pre-training}}\\
GCOPE~\cite{zhao2024all} & 44.29 ± 7.23 & 39.10 ± 8.39 & 50.82 ± 8.10 & 61.83 ± 6.48 & 51.12 ± 7.13 & 5.2\\
MDGPT~\cite{yu2024text} & 49.11 ± 7.63 & 43.56 ± 8.25 & 53.90 ± 9.83 & 63.41 ± 6.70 & 49.56 ± 7.63 & 3.8\\
BRIDGE~\cite{yuan2025much} & \underline{53.84 ± 7.52} & \textbf{50.49 ± 9.09} & \textbf{58.41 ± 8.52} & \underline{68.33 ± 5.65} & \textbf{54.06 ± 7.55} & \textbf{1.4}\\
\midrule
\textbf{\name~}(ours) &  \textbf{{62.31 ± 16.70}} & {43.93 ± 11.21}	& {52.40 ± 10.18} & \textbf{69.92 ± 12.27} & \underline{53.84 ± 11.42} & \underline{2.2}\\
\bottomrule
\end{tabular}}
\end{table*}

\subsection{Additional Experimental Results}
In this section, we provide addational experiment results.

\noindent\textbf{1-shot classification results} 
As shown in Table~\ref{tab:one_shot_node_cls}, \name~achieves the best average rank of 1.2 under the one-shot node classification setting, outperforming the strongest baseline BRIDGE with an average rank of 1.8. Specifically, \name~obtains the best performance on four out of five datasets, including Cora, CiteSeer, Photo, and Computers. The improvements are particularly significant on Cora and Photo, where \name~surpasses BRIDGE by 9.99$\%$ and 11.65$\%$, respectively. As shown in Table~\ref{tab:one_shot_graph_cls}, \name~achieves the best result on Cora and Photo with a clear improvement over BRIDGE. On the other datasets, BRIDGE remains the strongest baseline, while \name~achieves a competitive average rank of 2.2. This suggests that SET-GFM is effective in the one-shot graph classification setting, but graph-level transfer is more challenging than node-level transfer due to the information loss introduced by graph-level pooling and larger structural variation across graphs.

\begin{figure}[t]
  \centering
  \begin{minipage}{0.9\columnwidth}
    \centering
    \begin{subfigure}[b]{0.3\linewidth}
      \centering
      \includegraphics[width=\textwidth]{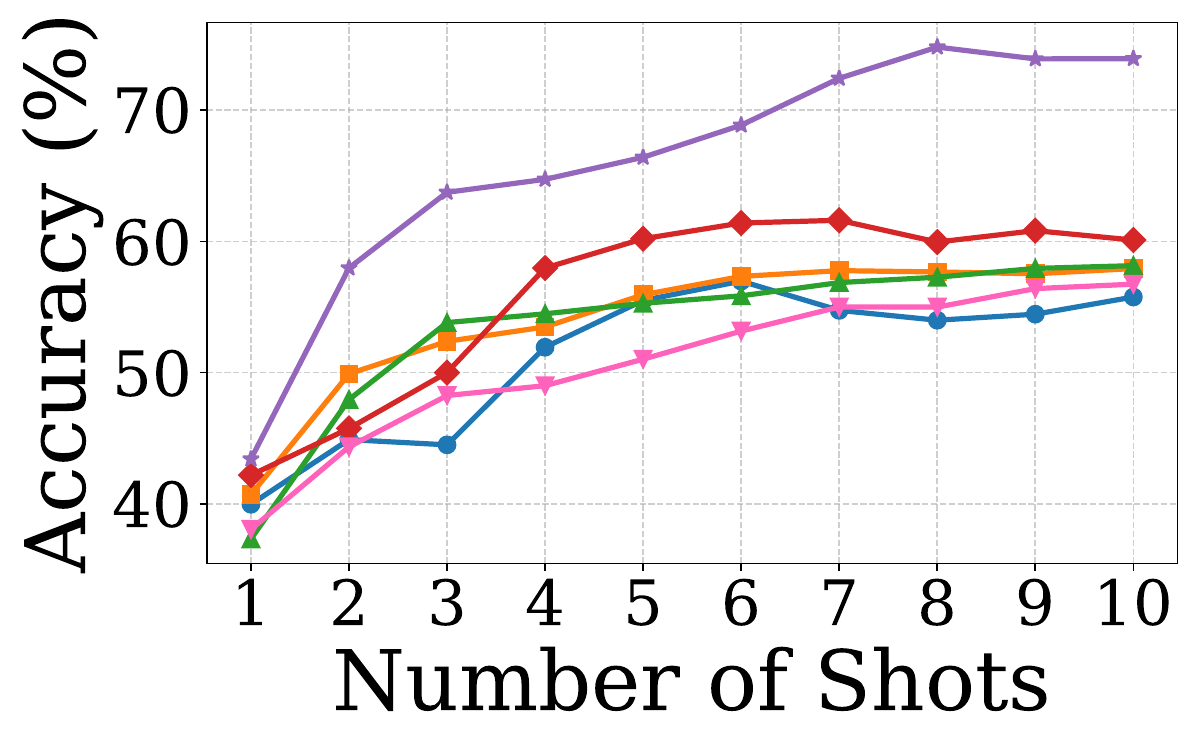}
      \caption{CiteSeer dataset}
      \label{fig:citeseer_node_appendix}
    \end{subfigure}%
    \hfill
    \begin{subfigure}[b]{0.3\linewidth}
      \centering
      \includegraphics[width=\textwidth]{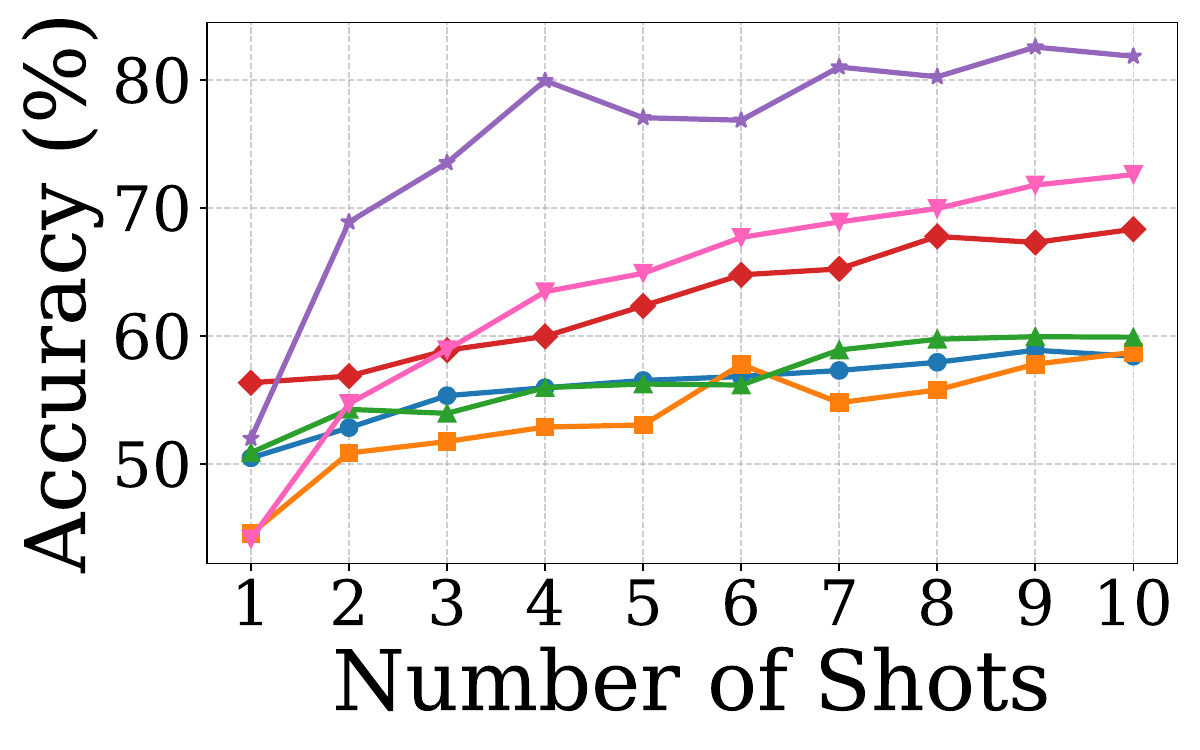}
      \caption{PubMed dataset}
      \label{fig:pubmed_node_appendix}
    \end{subfigure}
    \hfill
    \begin{subfigure}[b]{0.3\linewidth}
      \centering
      \includegraphics[width=\textwidth]{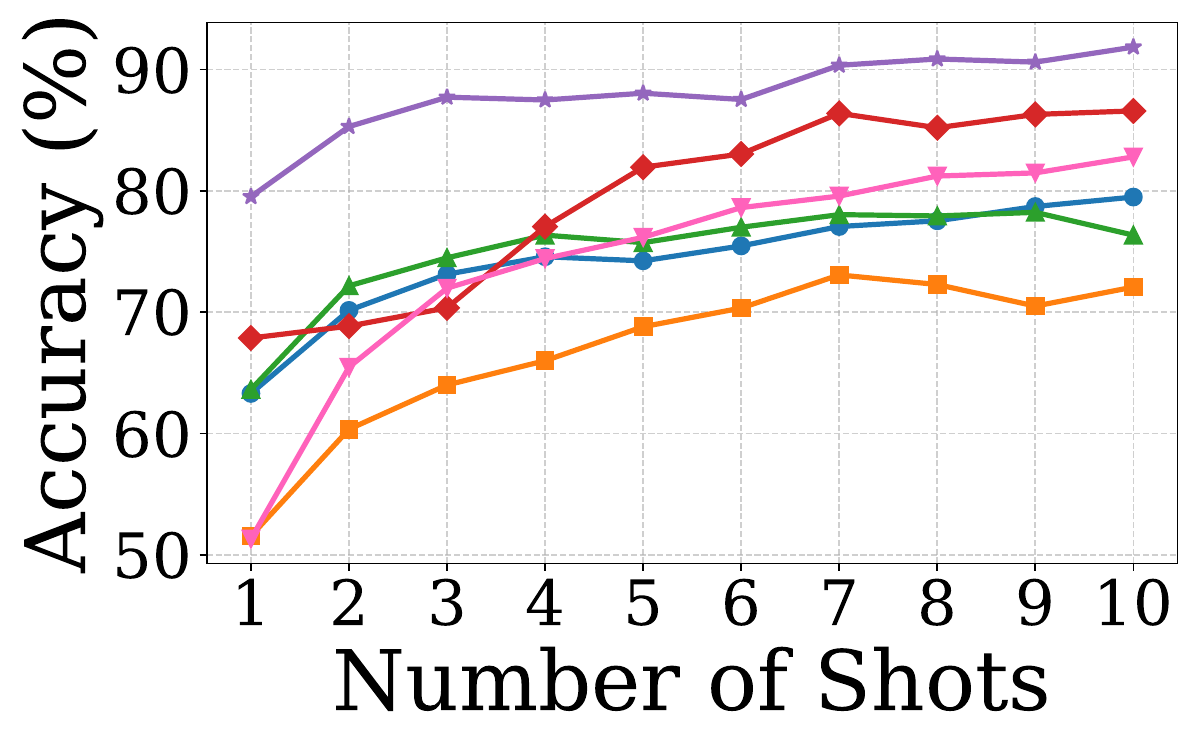}
      \caption{Photo dataset}
      \label{fig:photo_node_appendix}
    \end{subfigure}
    \medskip
    \includegraphics[width=0.98\linewidth]{fig/m-shot/methods_legend_node.pdf}
    \caption{$m$-shot performance on the target domain for node classification.}
    \label{fig:m-shot_node_cls_appendix}
  \end{minipage}%
\end{figure}

\begin{figure}[ht]
  \centering
  \begin{minipage}{0.9\columnwidth}
    \centering
    \begin{subfigure}[b]{0.3\linewidth}
      \centering
      \includegraphics[width=\textwidth]{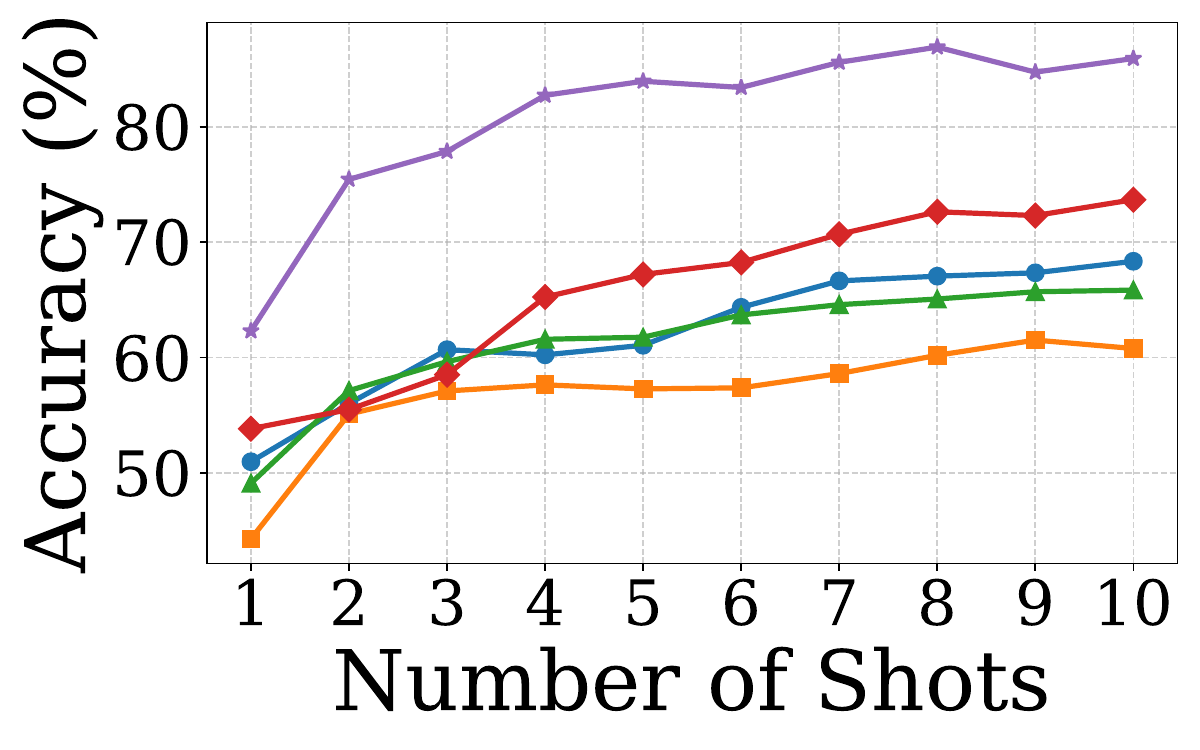}
      \caption{Cora dataset}
      \label{fig:cora_graph_appendix}
    \end{subfigure}%
    \hfill
    \begin{subfigure}[b]{0.3\linewidth}
      \centering
      \includegraphics[width=\textwidth]{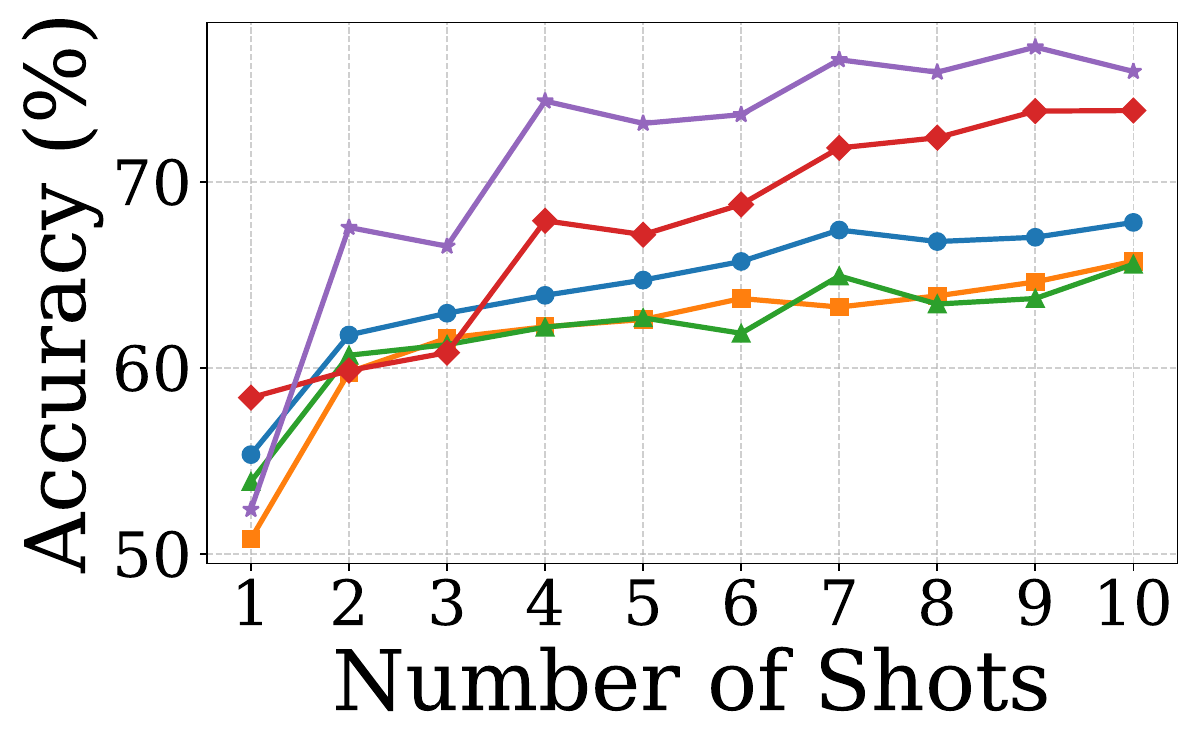}
      \caption{PubMed dataset}
      \label{fig:pubmed_graph_appendix}
    \end{subfigure}
    \hfill
    \begin{subfigure}[b]{0.3\linewidth}
      \centering
      \includegraphics[width=\textwidth]{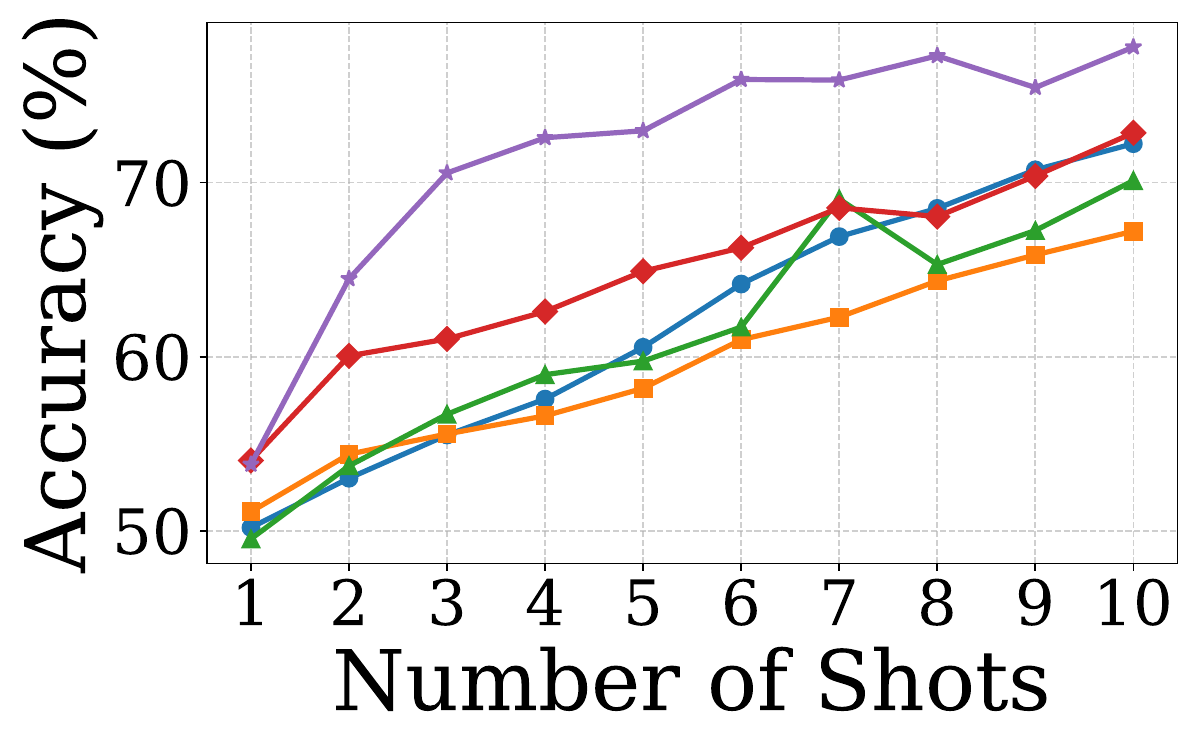}
      \caption{Computers dataset}
      \label{fig:compters_graph_appendix}
    \end{subfigure}
    \medskip
    \includegraphics[width=0.85\linewidth]{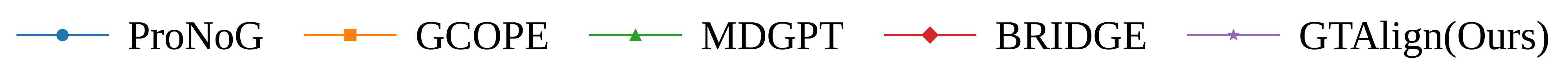}
    \caption{$m$-shot performance on the target domain for graph classification.}
    \label{fig:m-shot_graph_cls_appendix}
  \end{minipage}%
\end{figure}

\noindent\textbf{Different few-shot scenarios}
We also provide \name~'s performance in different few-shot scenarios, as shown in Figure~\ref{fig:m-shot_node_cls_appendix} and Figure~\ref{fig:m-shot_graph_cls_appendix}. For both node and graph classification, it can be observed that although the accuracy of our method may not be high at 1-shot, as the number of contextual samples increases, our method can effectively surpass existing methods and obtain more accurate predictions.

\begin{table*}[!t]
\caption{The results of ablation studies for five-shot graph classification. Reported results are Accuracy ± standard deviation for one hundred runs (\%). The best results are shown in \textbf{bold}.}
\label{tab:ablation_five_shot_graph}
\centering
\resizebox{\textwidth}{!}{%
\begin{tabular}{lccccc}
\toprule
Model / Target  & Cora & CiteSeer & PubMed & Photo & Computers \\
\midrule
\textbf{\name~}(full) &  \textbf{83.95 ± 5.39}	& \textbf{64.49 ± 7.21}	& \textbf{73.15 ± 7.08}	& \textbf{81.79 ± 4.66}	& \textbf{72.97 ± 6.30}\\
w/o GNN pre-train & 81.95 ± 5.64	&62.90 ± 6.54	&71.42 ± 7.40	&81.57 ± 5.05	 &71.53 ± 6.05 \\
w/o Continual pre-train &78.50 ± 7.77	&61.41 ± 7.66	&70.53 ± 8.00	 &78.23 ± 5.76	&71.98± 6.63\\
w/o GNN fine-tune & 66.10 ± 8.90	&50.83 ± 6.22	&60.66 ± 6.71	&76.88 ± 4.19	&65.33 ± 7.37\\
\bottomrule
\end{tabular}
}

\end{table*}

\begin{table*}[t]
\caption{The results of ablation studies for five-shot node classification. Reported results are Accuracy ± standard deviation for one hundred runs (\%). The best results are shown in \textbf{bold}.}
\label{tab:ablation_five_shot_node}
\centering
\resizebox{\textwidth}{!}{%
\begin{tabular}{lccccc}
\toprule
Model / Target  & Cora & CiteSeer & PubMed & Photo & Computers \\
\midrule
\textbf{\name~}(full)  & \textbf{88.65 ± 5.54}	&\textbf{66.41 ± 6.65}	&\textbf{77.05 ± 6.80}	&\textbf{87.64 ± 3.32}	&\textbf{82.94 ± 4.78} \\
w/o GNN pre-train & 86.06 ± 4.48	&65.44 ± 7.24	&73.53 ± 8.50	& 86.44 ± 3.63  &82.08 ± 5.21 \\
w/o Continual pre-train &82.29 ± 7.69	&64.64 ± 7.25	&73.95 ± 7.65	&83.83 ± 4.07	&81.11 ± 5.93\\
w/o GNN fine-tune  &57.62 ± 8.32	&50.42 ± 6.50 	&61.12 ± 7.85	&78.36 ± 4.12	 &67.30 ± 6.26\\
\bottomrule
\end{tabular}
}

\end{table*}

\subsection{Ablation Studies}
We also provide specific numerical tables about ablation studies across three key components of our framework, as shown in Table~\ref{tab:ablation_five_shot_graph} and Table~\ref{tab:ablation_five_shot_node}.
\begin{itemize}[nosep, leftmargin=*] 
    \item w/o GNN pre-train: We use a randomly initialized GNN encoder rather than a pre-trained one on multi-source domains.
    \item w/o Continual pre-train: We remove the second stage.
    \item w/o GNN fine-tune: We use directly use the pre-trained encoder weights without fine-tuning them using a few target samples in the third stage.
\end{itemize}

\begin{table*}[!t]
\caption{The results (Accuracy ± standard deviation for one hundred runs, $\%$) of \name~with different TFMs for five-shot node classification. The best results are shown in \textbf{bold}.}
\label{tab:different_tfm}
\begin{center}
\resizebox{\textwidth}{!}{
\begin{tabular}{lccccc}
\toprule
Model / Target  & Cora & CiteSeer & PubMed & Photo & Computers \\
\midrule
\textbf{\name~} + TabPFNv2.5 & 88.52 ± 5.52	&\textbf{67.59 ± 6.47}	&\textbf{77.34 ± 7.99}	&\textbf{87.76 ± 3.85}	&82.41 ± 5.05 \\
\textbf{\name~} + Limix-16M   & \textbf{88.65 ± 5.54}	&66.41 ± 6.65 & 77.05 ± 6.80	& 87.64 ± 3.32	&\textbf{82.94 ± 4.78} \\ 
\bottomrule 
\end{tabular}}
\end{center}
\end{table*}

\subsection{Different TFM Backbones}
We further evaluate \name~with different TFMs, TabPFNv2.5~\cite{grinsztajn2025tabpfn} and Limix-16M~\cite{zhang2025limix}, as shown in Table~\ref{tab:different_tfm}. This indicates that \name~is not sensitive to the choice of TFM, and the framework can consistently deliver strong performance regardless of whether TabPFNv2.5 or the more parameterized Limix-16M is used.

\end{document}